\documentclass{article}

\PassOptionsToPackage{numbers, sort&compress}{natbib}
 \usepackage[preprint]{neurips_2026}

\usepackage[utf8]{inputenc} 
\usepackage[T1]{fontenc}    
\usepackage{hyperref}       
\usepackage{url}            
\usepackage{booktabs}       
\usepackage{amsfonts}       
\usepackage{nicefrac}       
\usepackage{microtype}      
\usepackage{xcolor}         

\usepackage{graphicx}
\usepackage{subcaption}   
\usepackage{amsmath}
\usepackage{amssymb}
\usepackage{placeins}
\usepackage{adjustbox}

\usepackage{tikz}
\usepackage{color}
\usepackage{comment}
\usepackage{xfrac}

\usepackage{soul}
\usepackage{bbm}
\usepackage[capitalize]{cleveref}
\usepackage{pifont}
\newcommand{\cmark}{\ding{51}}
\newcommand{\xmark}{\ding{55}}

\newcommand*{\ShowNotes}{}
\definecolor{darkred}{rgb}{0.7,0.1,0.1}
\definecolor{darkgreen}{rgb}{0.1,0.7,0.1}
\definecolor{dblue}{rgb}{0.2,0.2,0.8}
\definecolor{maroon}{rgb}{0.76,.13,.28}
\definecolor{burntorange}{rgb}{0.81,.33,0}
\definecolor{cyan}{rgb}{0.0,0.7,0.94}
\definecolor{salmon}{rgb}{0.99,0.51,0.46}
\definecolor{green}{rgb}{0.03,0.91,0.43}
\definecolor{forestgreen}{rgb}{0.13,0.55,0.13}
\definecolor{grey}{rgb}{0.4,0.4,0.4}
\definecolor{purple}{rgb}{0.29,0,0.51}
\definecolor{crimson}{rgb}{0.86,0.08,0.24}

\ifdefined\ShowNotes
  \newcommand{\colornote}[3]{{\color{#1}\bf{#2: #3}\normalfont}}
\else
  \newcommand{\colornote}[3]{}
\fi

\usepackage[table]{xcolor}
\definecolor{softblue}{RGB}{220,230,250}

\usepackage{multirow}
\usepackage{stfloats}

\RequirePackage{xspace}
\makeatletter
\DeclareRobustCommand\onedot{\futurelet\@let@token\@onedot}
\def\@onedot{\ifx\@let@token.\else.\null\fi\xspace}

\def\eg{\emph{e.g}\onedot} 

\def\ie{\emph{i.e}\onedot}

\makeatother

\title{HOLA: Holistic Multi-Modal Alignment for Open-Set 3D Recognition}

\author{
  Koby Aharonov \\
  Technion – Israel Institute of Technology \\ 
  \texttt{koby\_a@campus.technion.ac.il} \\
  \And
  Oren Shrout \\
  Technion – Israel Institute of Technology \\ 
  \texttt{shrout.oren@campus.technion.ac.il} \\
  \And
  Ayellet Tal \\
  Technion - Israel Institute of Technology \\ 
  \texttt{ayellet@ee.technion.ac.il} \\
}

\begin{document}

\maketitle

\begin{abstract}
Open-set 3D recognition requires models that generalize to rare or unseen categories. Recent approaches address this by distilling language-vision knowledge into 3D encoders, typically relying on heavy 2D ViTs and aligning each point cloud with a single image or caption, thus anchoring representations to partial views. We propose aligning each point cloud with multiple images and textual descriptions to capture a more holistic understanding of 3D objects. To realize this idea, it is essential to design a loss function capable of jointly aligning a 3D instance with multiple matched signals, multi-view images and multiple texts, while separating positive aggregation from negative competition. We introduce such a function, termed the \emph{decoupled multi-positive contrastive loss}. Our formulation enhances the loss’s hardness-aware focus on challenging negatives, avoiding the “spotlight crowding” that occurs when many positives share the same softmax with all the negatives. Complementing this, we present a lightweight text adapter applied only to web captions, reducing the domain gap to curated annotations and enabling effective use of large-scale unsupervised text. Our model demonstrates state-of-the-art open-vocabulary performance on long-tail benchmarks, yielding substantial zero-shot improvements while sustaining high frame rates.
\end{abstract}

\section{Introduction}
Despite growing repositories of CAD models \cite{wu20153d,deitke2023objaverse,chang2015shapenet} and real-world scans \cite{uy2019revisiting,yu2023mvimgnet,song2015sun}, collecting and labeling high-quality 3D data remains slow, expensive, and fragmented across capture modalities (LiDAR, depth sensors, multi-view photogrammetry), each with its own artifacts and sparsity patterns. As a result, benchmarks over-represent a small set of “head” (frequent) categories while the “long-tail”, rare or domain-specific objects, appear infrequently or not at all. This imbalance collides with how 3D systems are deployed: robots, AR headsets, and autonomous platforms routinely encounter novel instances, subcategories, or entirely unseen classes in cluttered, partially observed scenes. In this setting, closed-set learning \cite{qi2017pointnet,qi2017pointnet++,wang2019dynamic,yu2022point,choy20194d} underperforms precisely because it expects complete label coverage.

To address the underrepresentation of many long-tail classes in labeled data, an open-set approach has been proposed \cite{Xue_2023_CVPR,NEURIPS2023_8c7304e7,zeng2023clip2}.
Concretely, a 3D encoder is trained to match CLIP’s \cite{radford2021learning, cherti2023reproducible} image-text shared embedding space, thereby transferring knowledge into 3D point-cloud embeddings.
These works jointly align heterogeneous signals: point clouds, multi-view renderings and textual descriptions, so that each modality contributes complementary cues about the same object. 
The result is a shared representation space that supports open-set recognition, enabling unknown 3D shapes to be named, retrieved, or recognized, thereby covering the long-tail categories in 3D.

Recent models demonstrate strong performance~\cite{zhou2023uni3d, Lei_2024_CVPR, qi2024shapellm}, but suffer from two main disadvantages.
First, they are computationally heavy and contain a large number of parameters (200M-2000M), which can lead to long training and inference times when computational resources are limited.
Second, the 3D-image/text alignment pipelines align a point-cloud feature with one image or one caption at a time, thereby anchoring the 3D representation to a partial view of the object rather than to its holistic shape. 
This limitation is particularly detrimental in the long tail, where 3D data is scarce and every available cue is essential.

Current efforts to remedy the latter issue attempt to leverage multiple rendered image views \cite{Zhang_2024_CVPR,Gao_2024_CVPR}.
While the results improve, these approaches still fall short of fully exploiting the richness of three-dimensional information.
This stems mostly from using the standard single-positive contrastive loss in an environment that seeks to incorporate multiple input signals.

\begin{figure}[t]
    \setlength\tabcolsep{3pt} 
    \centering
    \begin{subfigure}[t]{0.45\linewidth}
    \centering
    \includegraphics[width=\linewidth,height=0.2\textheight]{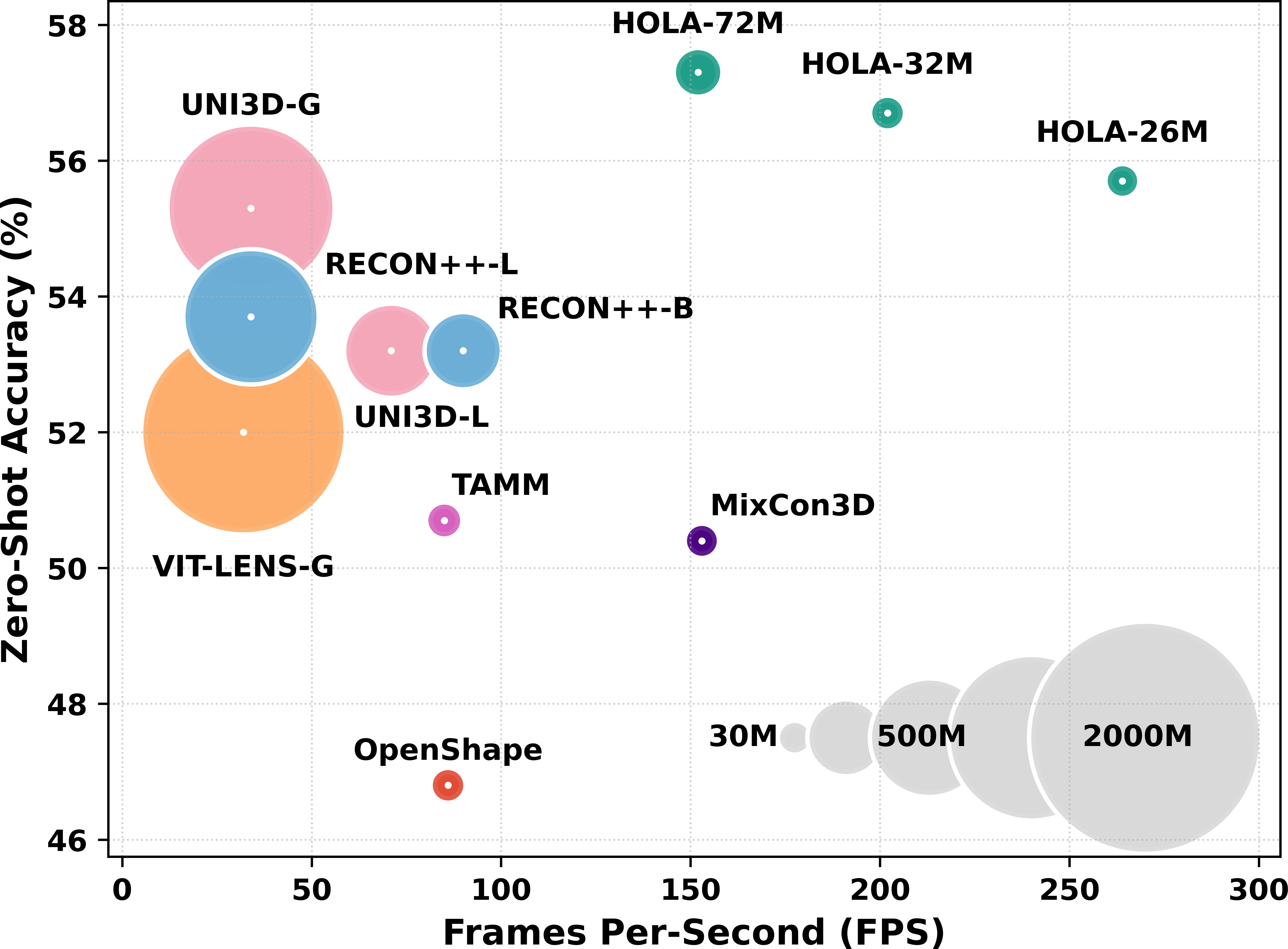}
    \caption{Performance vs. FPS.}
    \label{fig:teaser_performance_vs_fps}
    \end{subfigure}
    \begin{subfigure}[t]{0.2625\linewidth}
    \centering
    \includegraphics[width=\linewidth,height=0.2\textheight]{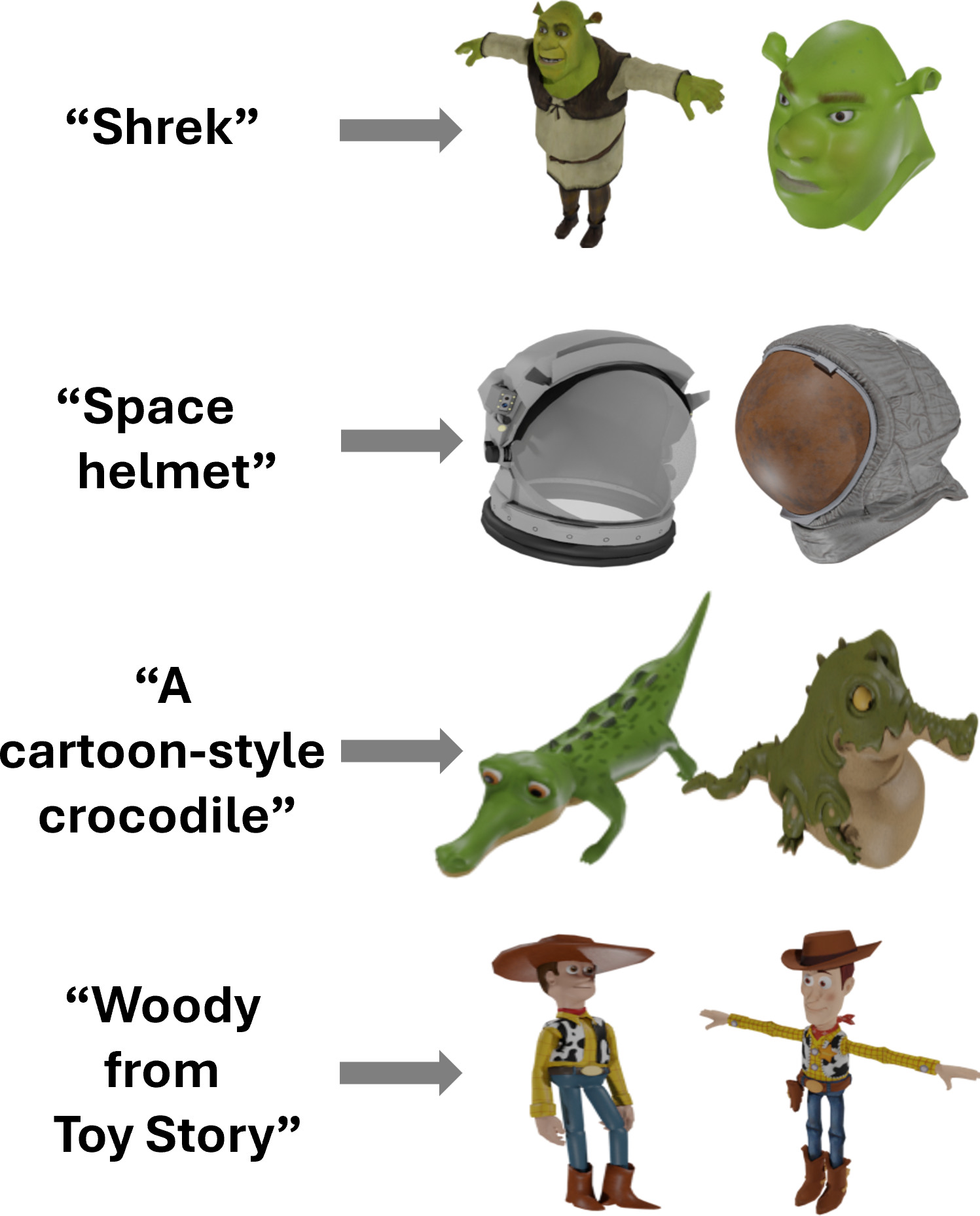}
    \caption{Text-to-3D retrieval}
    \label{fig:teaser_text_3d_ret}
    \end{subfigure}
    \begin{subfigure}[t]{0.2625\linewidth}
    \centering
    \includegraphics[width=\linewidth,height=0.2\textheight]{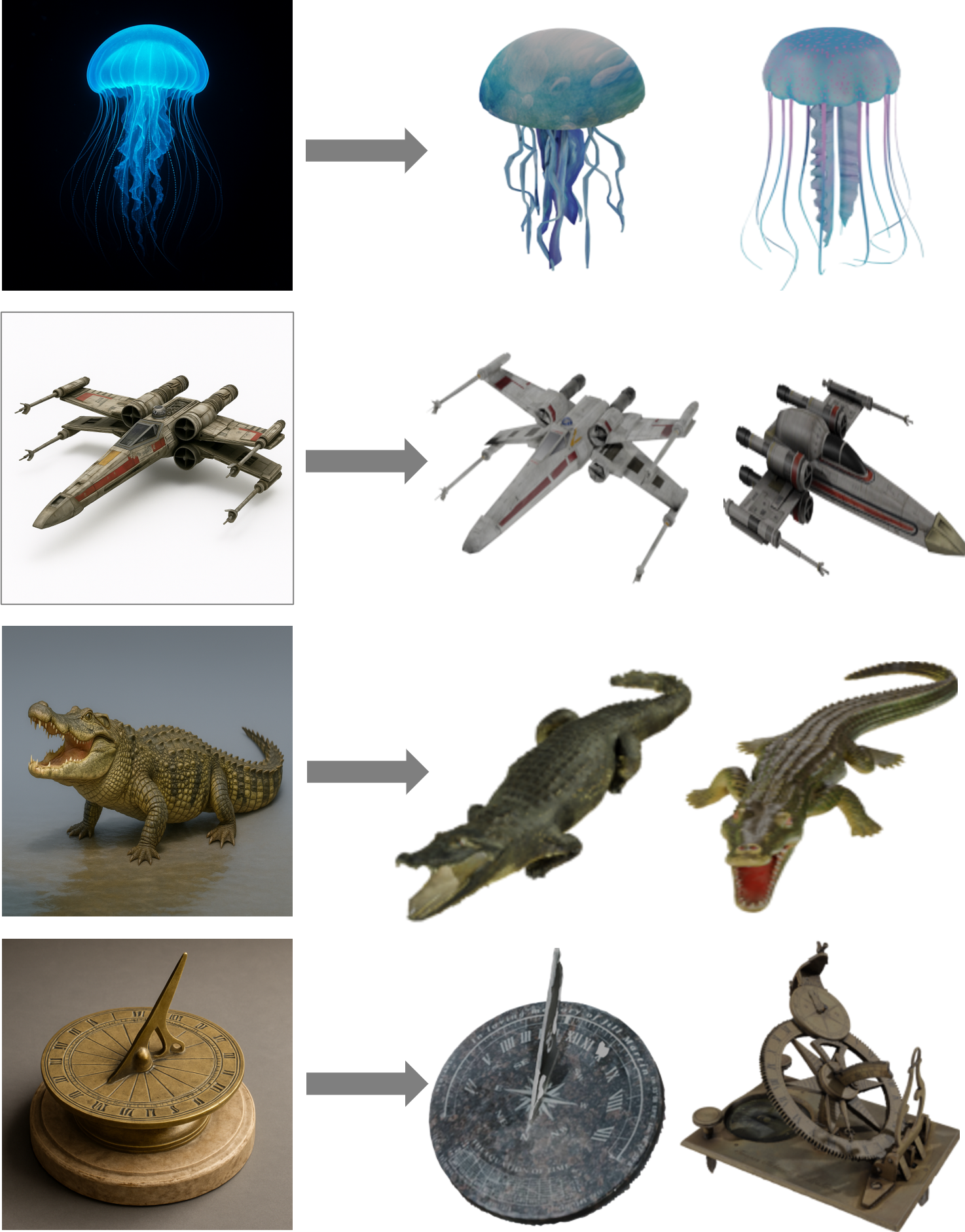}
    \caption{Image-to-3D retrieval}
    \label{fig:teaser_image_3d_ret}
    \end{subfigure}
    \caption{
    {\bf (a)}
    The plot shows that our models (in teal) achieve higher accuracy (vertical axis) and frame rates (horizontal axis) than competing methods, while using significantly fewer parameters (indicated by the smaller circle sizes).
    Results are shown for the challenging long-tail Objaverse-LVIS benchmark.
    {\bf (b) - (c)}
    Given text or image queries, our 3D shape retrieval results indicate a well-structured embedding space across tail categories. 
    }
    \label{fig:teaser}
\end{figure}

We introduce a new loss function designed to jointly embed multiple signals. 
We show that naively extending standard contrastive learning to a multi-positive contrastive objective, which allows multiple positive pairs per anchor sample~\cite{khosla2020supervised,tian2023stablerep}, is insufficient for this purpose.
Instead, our loss properly decouples the objective, enabling effective multi-signal learning. Using our loss function, even relatively lightweight models (\eg, 32M parameters) achieve state-of-the-art results on large, long-tail datasets (see \cref{fig:teaser}).

Specifically, one way to leverage the available views and modalities within a multi-positive contrastive objective is to contrast each point cloud with its corresponding views across modalities and then average the resulting similarities. 
However, naively adding positive views introduces a counterintuitive failure mode: as the number of positive views increases, the contrastive loss’s {\em hardness-aware property,} that is, its ability to focus on informative negative samples, gradually weakens.
This occurs because the naive form of the loss regards all examples (both positives and negatives) as a single set. It selects the most informative samples to attract or repel based on their similarity to the anchor. Consequently, positive views are more likely to be chosen, leaving the negatives less attended.

Mathematically, the softmax-based contrastive loss is hardness-aware as its gradients place greater weight on hard negatives (contrastive samples).
Adding many positive views per point cloud can inadvertently blunt this effect: probability mass shifts toward positives, so gradients from hard negatives are down-weighted, reducing the most informative separations. 
However, these hard negatives are crucial, as they are the most confusing for the model, being the samples most similar to the anchor point cloud.
We will show that decoupling the contrastive loss mitigates this by separating (i) a multi-positive alignment term from (ii) a negative-separation term, thereby preserving and even amplifying hardness-aware property while still aggregating complementary cues across views and modalities. 
This effect is especially important in long-tail data, where classes contain very few examples. In such cases, the additional information provided by multiple views becomes more valuable, both for enriching the representation of rare classes and for improving their separation from negative examples.

Additionally, we observe a domain gap among text sources, in particular between in-domain annotations of rendered views and retrieved real-image captions describing the same 3D object.
During training, a text source is randomly sampled for each shape.
To address this, we propose a lightweight MLP adapter applied exclusively to retrieved-text embeddings, aligning them with in-domain annotations and reducing cross-domain drift.

We realized the above ideas using two relatively small 3D backbones: PointBERT \cite{yu2022point} and MinkowskiNet \cite{choy20194d}.
Like all recent approaches, we distill features from frozen encoders taken from CLIP \cite{radford2021learning,cherti2023reproducible}.
We demonstrate state-of-the-art zero-shot performance on large, long-tail datasets across most training protocols.
Our model uses substantially fewer parameters and achieves significantly faster inference times compared to the largest state-of-the-art models.

In summary, our main contributions are as follows:
\begin{itemize}
    \item
    We propose a new loss function that jointly aligns 3D shapes with their corresponding image and text views.
    By decoupling multi-positive alignment from negative separation, it amplifies the loss’s hardness-aware behavior.    
    \item
    We introduce a lightweight adapter for the captions of retrieved real images, which mitigates the domain gap from object annotations and enables effective use of the richness of real-image captions data.
    \item
    Our method advances open-set recognition and long-tail generalization while retaining lightweight computation. 
    On the long-tailed Objaverse-LVIS benchmark, we observe a $2.0\%$ improvement in zero-shot performance, surpassing methods that use the strongest 2D models (1020M parameters), despite employing a much smaller model (72M). 
    We also outperform the best lightweight baseline (35M) by $6.0\%$ using a model of comparable size. 
\end{itemize}

\section{Related work}
\noindent
{\bf Multi-modal 3D representation learning.}
The goal of multi-modal 3D representation learning is to learn a unified feature space for 3D data by leveraging complementary information from multiple modalities.
A parallel line of research, including PointCLIP~\cite{Zhang_2022_CVPR, Zhu_2023_ICCV}, CLIP2Point~\cite{huang2023clip2point}, and Duoduo CLIP~\cite{lee2025duoduo}, uses 2D vision priors with rendered views or depth maps of point clouds, thereby operating in the image domain rather than directly on raw point clouds. 

Recent work that focuses directly on point clouds has largely evolved along two directions. 
The first consists of two-modality approaches that align point clouds with images, such as CrossPoint~\cite{afham2022crosspoint} and PiMAE~\cite{chen2023pimae}. 
A second direction comprises three-modality methods that distill knowledge from vision-language priors~\cite{radford2021learning,cherti2023reproducible,mu2022slip}, such as ULIP~\cite{Xue_2023_CVPR, Xue_2024_CVPR}, OpenShape~\cite{NEURIPS2023_8c7304e7}, and CLIP$^2$~\cite{zeng2023clip2}, which have substantially advanced zero-shot transfer and cross-modal retrieval.

Building on the latter, subsequent point cloud based methods have broadly followed two design trends. 
One trend retains relatively compact 3D backbones, as in TAMM \cite{Zhang_2024_CVPR} and MIXCON3D \cite{Gao_2024_CVPR}, with an emphasis on lightweight models and efficient training and inference. 
The other adopts larger ViT-based encoders, including UNI3D \cite{zhou2023uni3d}, ViT-Lens \cite{Lei_2024_CVPR}, and ReCon++ \cite{qi2024shapellm}, increasing model capacity to strengthen cross-modal alignment and improve representation quality. 
We follow the former direction, favoring small, efficient backbones while focusing on how to leverage multiple textual shape descriptions and image views.

\noindent
{\bf Contrastive loss.}
Contrastive learning is a self-supervised approach that trains models to bring related samples closer and push unrelated ones apart \cite{oord2018representation,chen2020simple,he2020momentum}. This framework has proven highly effective and has inspired a broad family of objectives and architectures; we refer readers to a recent survey for a comprehensive overview~\cite{hu2024comprehensive}.

The effectiveness of contrastive learning can degrade when batch sizes are small, as the likelihood of sampling hard negatives decreases.
Several works propose remedies, such as hard negative mining \cite{kalantidis2020hard,robinson2020contrastive}, a reweighting strategy \cite{tsai2021self}, or decoupled contrastive learning in InfoNCE loss \cite{yeh2022decoupled}. 
The latter has been applied to a range of tasks, particularly medical segmentation~\cite{koleilat2025medclip} and classification~\cite{qiu2024learn}.

In our setup, which involves multiple views, a key challenge is that the contribution of hard negatives diminishes regardless of batch size.
We analyze this issue and propose a decoupled multi-positive contrastive loss tailored to multi-modal alignment, ensuring strong positive cohesion while enforcing meaningful separation from negative samples.

\begin{figure}[t]
    \setlength\tabcolsep{3pt} 
    \centering
    \includegraphics[width=0.95\linewidth,height=0.2\textheight]{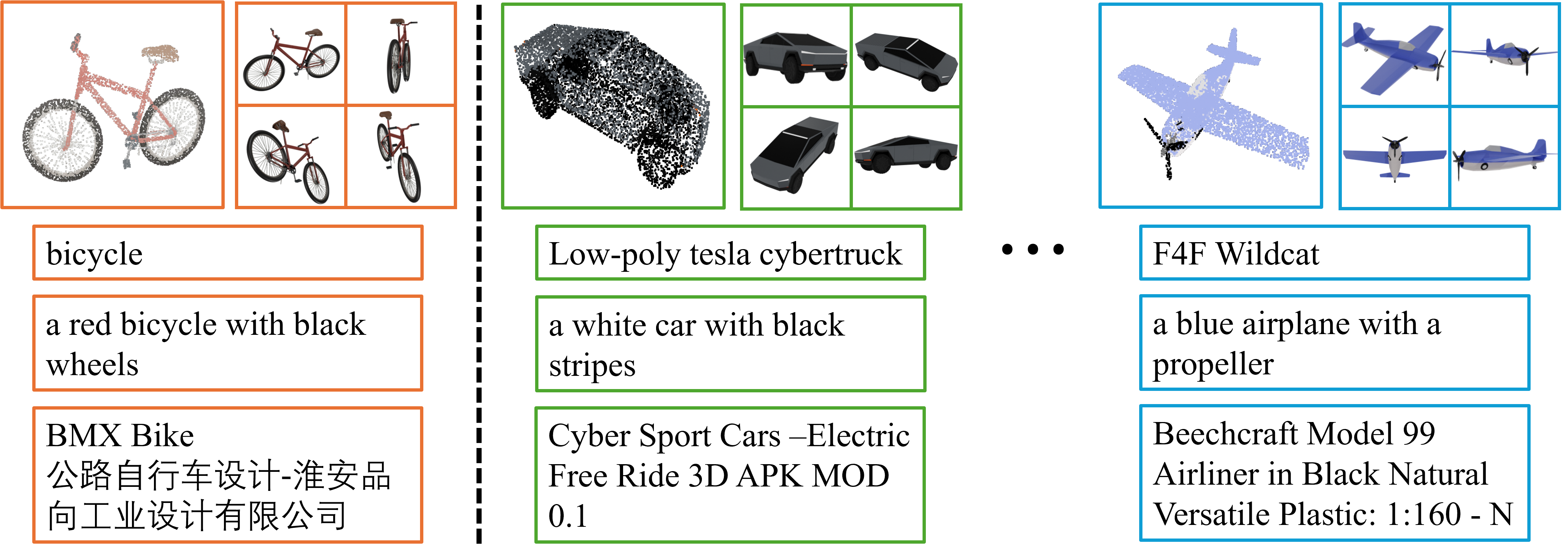}\\
    \hspace{-0.125in} (a) Anchor triplet \hspace{0.75in}  ($b_1$) Triplet $1$ \hspace{0.51in} {\bf $\dots$}  \hspace{0.39in} ($b_{N}$) Triplet $N$
    \caption{
    {\bf Input Triplets.} 
    Each training triplet consists of a point cloud, a set of rendered images, and a set of texts. The texts may include (from top to bottom): (i) annotations, (ii) VLM-generated captions, and (iii) retrieved web captions. 
    During training, the modalities within each triplet are used in turn as anchors: when one modality serves as the anchor, the remaining matched modalities from the same triplet serve as positive keys ($a$). 
    Modalities from the other triplets in the batch serve as negative samples ($b_{1},\dots,b_{N}$).
    }
    \label{fig:input}
\end{figure}

\section{Method}

Given a point cloud, we associate it with a corresponding set of image views and multiple text descriptions, and treat text as an additional "view" analogous to images (\cref{fig:input}).
The goal is to learn a 3D embedding of the point cloud that is aligned with the embeddings of its corresponding text and image views.
This joint alignment enables open-vocabulary understanding, equipping the model for open-set recognition tasks such as zero-shot classification and retrieval.

The input set of images consists of renderings of the object from multiple viewpoints. 
The associated textual descriptions can be of three types:
(i) human-provided annotations or metadata, when available;
(ii) automatically generated captions of the rendered images, produced by models such as BLIP \cite{li2022blip} or Azure; and
(iii) captions of retrieved real-world images (\eg, from LAION-5B \cite{schuhmann2022laion}), obtained by querying the dataset with a rendered image.
While the first type offers highly reliable descriptions, it is rarely available. 
The second type is generally of moderate to good quality, depending on the chosen viewpoint and captioning model. 
The third type is less reliable, as the retrieved image may not precisely correspond to the rendered object, leading to captions that only approximately describe the point cloud.

The primary challenge is to develop a tri-modal, symmetric alignment that encourages the learned 3D embeddings to be jointly consistent with both text and image features, while also preserving cross-consistency between text and images.
To this end, we propose a one-to-many loss formulation (\eg, one point cloud and multiple images), which naturally extends to a many-to-many setting in the case of image-text alignment.
Therefore, much of this section will be devoted to the formulation of a reliable loss function between modalities.

\begin{figure*}[t]
    \setlength\tabcolsep{3pt} 
    \centering
    \includegraphics[width=\linewidth,height=0.18\textheight]{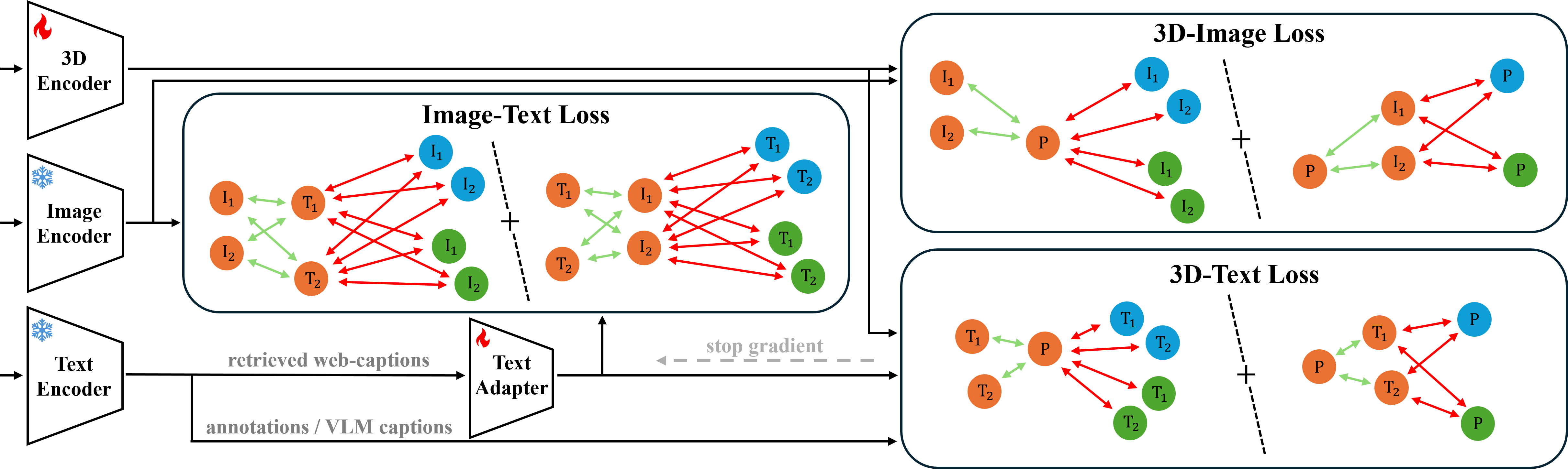} 
    \caption{
    {\bf Model.} After each modality is encoded independently, pairwise losses are computed across all modality pairs. The decoupled multi-positive contrastive loss (shown in each rectangle) is one-to-many when involving point clouds and many-to-many when aligning images and texts.
    The image and text encoders are kept frozen, while the 3D encoder and the text adapter are trainable.
     Note that the colors (orange, cyan, and green) correspond to those in \cref{fig:input}; specifically, the anchor and positive keys are from the modalities of the orange training triplet, while the other samples (across all modalities), shown in cyan and green, serve as negative keys.
    }
    \label{fig:model}
\end{figure*}

Once the loss function is established, the overall model design follows naturally, as illustrated in \cref{fig:model}.
It comprises three encoders, one for each modality.
The image and text encoders are kept frozen, while the 3D encoder is trainable.
In addition, to reduce the domain gap between retrieved web captions and the in-domain text sources, we introduce a lightweight trainable text adapter. 
The adapter is applied exclusively to retrieved web-caption embeddings, enabling the model to better exploit their linguistic diversity and world knowledge.
In our implementation, we adopt frozen CLIP-style \cite{radford2021learning,cherti2023reproducible} text and image encoders pretrained on web-scale 2D corpora, capitalizing on their large-scale pretraining.

In \cref{subsec:DecoupledConLoss}, we formalize the training objective, deriving the multi-positive decoupled loss and showing how it preserves hard-negative focus even when each shape has many positive views.
In \cref{subsec:Text_source_alignment}, we introduce the lightweight adapter for handling retrieved web captions.

\subsection{Loss function}
\label{subsec:DecoupledConLoss}
Our goal is to design a loss function that
(1) effectively handles multi-positive inputs (\ie, multiple images or text descriptions) by aligning all of them with the corresponding point cloud anchor;
(2)~robustly addresses classes in the long-tail; and
(3) enables the use of a compact backbone with a small number of parameters, ensuring computational efficiency and fast inference.

The conventional single-positive contrastive loss is ill-suited to our setting, as it fails to account for multiple views, an aspect particularly important in long-tail scenarios.
However, simply extending it to a naive multi-positive formulation is also inadequate, as increasing the number of positives can degrade performance by 
reducing the emphasis on negative-sample gradients, thereby weakening the hardness-aware property.
To address this issue, our key idea is to align all positive samples for each instance (multi-positive alignment) while decoupling this process from the negative separation term that repels non-matching samples.
Accordingly, we propose a multi-positive loss that ensures effective treatment of negative samples without interfering with positives.
We further extend this formulation by introducing hardness-aware weighting for positive samples, allowing the model to assign greater importance to more challenging positives during training.
In this loss, the gradients are more stable and informative, leading to faster and more reliable convergence, which in turn enables the use of compact backbones.
The details are discussed below.

\noindent
{\bf Single-positive contrastive loss.}
In our setting, each point cloud is associated with multiple matching views. Before explaining how we handle multiple positives, we first review the standard loss formulation that considers only a single positive sample at a time, ignoring the others \cite{oord2018representation,chen2020simple,he2020momentum}.
Thus, its goal is to attract the anchor (\eg query point cloud) closer to its single-positive matching sample (\eg view) while it is repelled away from many non-matching negative samples. 

We formalize the loss function to enable its later generalization to multi-positive views.
Given an anchor (query) embedding \(q_i\), its positive key embedding  \(k^{\scriptscriptstyle +}\),  and its negative keys \(\{k^{\scriptscriptstyle -}_{j}\}_{j=1}^{N}\), the  similarity scores between the \(\ell_2\)-normalized anchor and key vectors are defined as:
$\quad s^{\scriptscriptstyle +}_{i}=q_i \cdot k^{\scriptscriptstyle +},\quad \big\{ s^{\scriptscriptstyle -}_{ij}=q_i \cdot k^{\scriptscriptstyle -}_{j}\big\}_{j=1}^{N}.$

The per-anchor single-positive contrastive loss is the negative log-likelihood of selecting the positive among all candidates, where \(\tau>0\) is a temperature:
\begin{equation}
\label{eq:single_positive_cl}
\mathcal{L}^{\textit{SP}}_{q_i}
= -\log
   \frac{\exp\left(s^{\scriptscriptstyle +}_{i}/\tau\right)}
        {\exp\left(s^{\scriptscriptstyle +}_{i}/\tau\right)
        +  \sum\limits_{j}^{} \exp\left(s^{\scriptscriptstyle -}_{ij} / \tau\right)} \;\;.
\end{equation}

\noindent
{\bf Multi-positive contrastive loss.}
Our goal is to align the 3D point-cloud embedding with all of its corresponding views.
This is particularly beneficial in scenarios with limited data or long-tail distributions, as utilizing multiple image/text positives promotes better generalization to rare or unseen categories.

The loss should simultaneously attract an anchor towards all of its positive samples (\ie, instances of the same class) while repelling it from all negative samples in the batch~\cite{khosla2020supervised, tian2023stablerep}. 
We define the similarity scores between the \(\ell_2\)-normalized anchor and key vectors as follows:
$
\big\{s^{\scriptscriptstyle +}_{ir}=q_i \cdot k^{\scriptscriptstyle +}_{r}\big\}_{r \in P(i)}, \big\{ s^{\scriptscriptstyle -}_{ij}=q_i \cdot k^{\scriptscriptstyle -}_{j}\big\}_{j \in N(i)},
$
where \(P(i)\) is the subset of indices of positive keys, \(N(i)\) is the subset of indices of negative keys, \(\{k^{\scriptscriptstyle +}_{r}\}_{r \in P(i)}\) are its positive key embeddings and \(\{k^{\scriptscriptstyle -}_{j}\}_{j \in N(i)}\) are its negative key embeddings.

We define the probability of a positive key $k^{\scriptscriptstyle +}_{r}$ being recognized as (a view of) anchor \(q_i\) as:
\begin{equation}
\label{eq:prob_of_pos_samples}
P^{\scriptscriptstyle +}_{ir}\triangleq
\frac{\exp\left(s^{\scriptscriptstyle +}_{ir}/\tau\right)}{ \sum\limits_{r^{\prime} \in P(i)}\exp\left(s^{\scriptscriptstyle +}_{ir^{\prime}}/\tau\right)\;\;+\sum\limits_{j \in N(i)}\exp\left(s^{\scriptscriptstyle -}_{ij}/\tau\right)}\;\;.
\end{equation}
Subsequently, the per-anchor multi-positive loss is the mean of negative log-likelihood terms of the samples recognized as the anchor:
\begin{equation}
\label{eq:multi_positive_cl}
\mathcal{L}^{\textit{MP}}_{q_i}=
 -\frac{1}{|P(i)|}\sum\limits_{r\in P(i)}\log P^{\scriptscriptstyle +}_{ir} \;\;.
\end{equation}


\noindent
{\bf Analyzing the hardness-aware property of \cref{eq:multi_positive_cl}.}
We observe that the multi-positive contrastive loss may fail when additional positive examples are introduced, as this violates the hardness-aware property \cite{wang2021understanding} that allows the loss function to focus on the most confusing samples (hard negatives) during training. 
We analyze this issue in detail below. 
This analysis is particularly important because our method relies on multiple positive examples; therefore, restoring this property in the loss function is essential for its effectiveness.

Mathematically, the hardness aware property of a loss function is reflected in the magnitude of its gradients w.r.t. negative samples.
Since the gradient magnitude is proportional to the similarity between the anchor and the negative sample, we expect that upon convergence, negative samples will be distant from the anchor and their associated gradient magnitudes will be small.
However, during training, the hardest negative samples typically produce large gradients, which serve to push them farther from the anchor. 
Unfortunately, when many positive samples are present, these hard negative samples can also become associated with small gradients, since all gradients are normalized jointly. This diminishes the loss’s ability to separate them effectively, thereby impairing the training process.
Formally, the gradients of $\mathcal{L}^{\textit{MP}}_{q_i}$ (\cref{eq:multi_positive_cl}) with respect to the positive/negative key similarities $s^{\scriptscriptstyle +}_{ir}$/$s^{\scriptscriptstyle -}_{ij}$ are: 
\begin{equation}
\label{eq:gradients}
\frac{\partial \mathcal{L}^{\textit{MP}}_{q_i}}{\partial s^{\scriptscriptstyle +}_{ir}}
=\frac{1}{\tau}\left[P^{\scriptscriptstyle +}_{ir} - \frac{1}{|P(i)|}\right]
\;,\;\;
\frac{\partial \mathcal{L}^{\textit{MP}}_{q_i}}{\partial s^{\scriptscriptstyle -}_{ij}}
=\frac{1}{\tau}P^{\scriptscriptstyle -}_{ij} \;\;,
\end{equation}
where $P^{\scriptscriptstyle -}_{ij}$ is the probability of a negative key $k^{\scriptscriptstyle -}_{j}$ being recognized as (a view of) anchor \(q_i\):
\begin{equation}
\label{eq:prob_of_neg_samples}
P^{\scriptscriptstyle -}_{ij}\triangleq
\frac{\exp\left(s^{\scriptscriptstyle -}_{ij}/\tau\right)}{ \sum\limits_{r \in P(i)}\exp\left(s^{\scriptscriptstyle +}_{ir}/\tau\right)\;\;+\sum\limits_{j^{\prime} \in N(i)}\exp\left(s^{\scriptscriptstyle -}_{ij^{\prime}}/\tau\right)}\;\;.
\end{equation}

\cref{eq:gradients,eq:prob_of_neg_samples} show that gradients w.r.t. negative key similarities $s^{\scriptscriptstyle-}_{ij}$ are proportional to \(exp(s^{\scriptscriptstyle-}_{ij}/ \tau)\).
Thus, hard negatives (those with larger $s^{\scriptscriptstyle-}_{ij}$ \ie, more similar to the anchor) receive larger gradient weights, w.r.t other negative gradients. 
This indicates that the multi-positive contrastive loss is indeed a hardness-aware loss: the gradient concentrates on the negatives nearest to (\ie., most confusable with) the anchor. 

However, from the denominator of \cref{eq:prob_of_neg_samples}, it can be observed that as more positive examples are added, the gradient magnitudes of the hard negatives decrease. 
The challenge, therefore, is to calibrate the objective so that the gradient magnitude reflects the sample's hardness even in the presence of many positive examples.
This is challenging because the positive and negative samples are inherently coupled, modifying one will inevitably affect the other.
The following equation, derived from \cref{eq:prob_of_pos_samples,eq:prob_of_neg_samples,eq:gradients}, shows the relationship between the gradients with respect to the positive and negative similarities:

{
\begin{equation}
\label{eq:pos_neg_gradients_relationship}
\sum\limits_{j \in N(i)} \frac{\partial \mathcal{L}^{\textit{MP}}_{q_i}}{\partial s^{\scriptscriptstyle -}_{ij}}
=
\frac{1}{\tau}\sum\limits_{j \in N(i)}P^{\scriptscriptstyle -}_{ij}
=
\frac{1}{\tau}\left[1 - \sum\limits_{r \in P(i)}P^{\scriptscriptstyle +}_{ir}\right] 
=
\left|\sum\limits_{r \in P(i)} \frac{\partial \mathcal{L}^{\textit{MP}}_{q_i}}{\partial s^{\scriptscriptstyle +}_{ir}}\right|
.
\end{equation}
}

This implies that the hardness-aware property also depends on the positive keys through the following term: 
$\sum_{r \in P(i)}P^{\scriptscriptstyle +}_{ir}$. 
As the positive keys get closer to the anchor (\ie, higher similarity and thus higher value for $\sum_{r \in P(i)}P^{\scriptscriptstyle +}_{ir}$), the loss allocates less weight to gradients w.r.t. the negative keys (\ie. smaller value for $\sum_{j \in N(i)}P^{\scriptscriptstyle -}_{ij}$) and the repulsive pressure from negative samples diminishes, meaning that the hardness-aware property weakens.

Because positive samples typically lie in close proximity to the anchor, they dominate the allocation of probability mass and, consequently, the gradients w.r.t. to the negative samples are less attended. 
Hard negatives, despite being the most informative, receive comparatively less attention and their contribution to the embedding space formation diminishes. 
Thus, the training signal is reallocated toward reinforcing already well-aligned positives rather than separating confusable negatives, attenuating the hardness-aware behavior of contrastive learning.

\begin{figure*}[t]
    \setlength\tabcolsep{3pt} 
    \centering
    \includegraphics[width=0.98\linewidth]{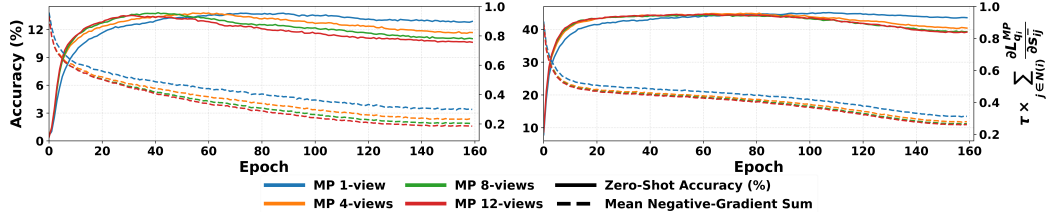} 
    \caption{
    {\bf Effect of the “spotlight crowding” phenomenon.}
    As views increase, the naive MP loss suppresses negative-sample gradients (dashed curves decline), leading to a larger drop in accuracy (solid curves).
    Results are reported on Objaverse-LVIS Top-1 accuracy for the "ShapeNet" (left) and "Ensembled no LVIS" (right) training sets.
    }
    \label{fig:spotlight_crowding}
\end{figure*}

\noindent
{\bf Decoupled multi-positive contrastive loss.}
We aim to resolve the aforementioned issue, wherein negative samples are insufficiently attended to and thus contribute less to the learning process despite their significance.
Mathematically, this issue is evident in the standard multi-positive formulation of \cref{eq:gradients,eq:prob_of_neg_samples}, where the terms corresponding to positive and negative samples compete for “spotlights”. 
As the number of positive samples increases (\ie, with multiple views), the positive terms receive disproportionately greater weight. 
Because positive examples are located close to the anchor, while only adversarial negatives lie nearby, the dominance of the positive terms continues to grow with the number of views.
This results in gradients being dominated by positive terms, making it difficult for the model to effectively learn from hard negatives.
We term this phenomenon “spotlight crowding”, and its ramifications can be seen in \cref{fig:spotlight_crowding}.

To address this issue, our key idea is to decouple the positive terms from the negative terms. This prevents the positive and negative terms from competing for “spotlights”. 
Instead, negative samples compete only with other negatives for gradient mass, ensuring that the most confusable samples receive the strongest corrective updates. 
As a result, the loss amplifies its hardness-aware behavior.

To this end, we introduce the \emph{decoupled multi-positive contrastive loss}, which directly addresses this problem, derived by omitting the denominator terms corresponding to positive key similarities in \cref{eq:prob_of_pos_samples,eq:multi_positive_cl}.
It is defined as: 
\begin{equation}
\label{eq:decoupled_mu_pos_con_loss}
\mathcal{L}^{\textit{DMP}}_{q_i}
=
-\dfrac{1}{\tau|P(i)|}\sum\limits_{r\in P(i)}s^{\scriptscriptstyle +}_{ir} \;+\; \log\sum\limits_{j\in N(i)} \exp\left(s^{\scriptscriptstyle -}_{ij}/\tau\right) \;\;,
\end{equation}
where \(\tau>0\) is a given temperature.
This loss consists of two complementary components with different objectives: the first enforces multi-positive alignment and the second is responsible for repelling non-matching objects. 

The first term aggregates all matched views of an instance, pulling them toward a single anchor to enforce cross-view consistency. 
This consolidation is particularly important for long-tail data, where each class may only have a few instances:   
pooling complementary cues from different views yields a more holistic and stable embedding, improving robustness to data scarcity and view bias.

The second term enforces separation by driving negative samples away from the anchor.
To understand its contribution, we examine the derivative of \cref{eq:decoupled_mu_pos_con_loss} with respect to the negative key similarities $s^{\scriptscriptstyle -}_{ij}$. 
It is defined as:
\begin{equation}
\label{eq:decoupled_negative_gradients}
\dfrac{\partial \mathcal{L}^{\textit{DMP}}_{q_i}}{\partial s^{\scriptscriptstyle -}_{ij}}
=\frac{1}{\tau}\dfrac{\exp\left(s^{\scriptscriptstyle -}_{ij}/\tau\right)}
{\sum\limits_{k \in N(i)} \exp\left(s^{\scriptscriptstyle -}_{ik}/\tau\right)} \;\;.
\end{equation}
The improved separation  occurs because the softmax denominator in \cref{eq:decoupled_negative_gradients} contains no terms corresponding to positive key similarities; therefore, the gradients with respect to the negative key similarities are unaffected by the number of positive samples. 
Consequently, the loss amplifies its hardness-aware behavior, as negative samples compete within themselves for gradient mass, ensuring that the hard negative samples receive the strongest updates (see \cref{supp_sec:analysis_of_hardness_aware_simple_loss} in the supplementary material for a detailed analysis of hardness-aware weighting).
This leads to better decision boundaries, and yields embeddings that better discriminate rare categories common in the long tail.

The loss structure also benefits small networks: by preventing gradient dilution from many positives and making negatives compete only with one another, the loss provides a sharper, higher-fidelity gradient signal. 
The result is more sample-efficient learning and cleaner decision boundaries without relying on large model capacity, allowing compact 3D backbones to capture holistic semantics and long-tail distinctions.
This is demonstrated empirically in \cref{sec:experiments}.

\noindent
\textbf{Weighting function for the multi-positive alignment term.}
We introduce a weighting function for the multi-positive alignment term that mirrors the negative term's hardness-aware property. 
As with the negative term, where learning benefits most from difficult (hard) negative samples, we prioritize hard anchors on the positive side: anchors that remain farther from their positive samples (average lower similarity) carry a stronger corrective signal and therefore receive larger weights, while easy anchors are down-weighted.

Formally, let $\mathcal{B}_{y_i} \;=\; \{\, l \in \mathcal{B} \mid y_l = y_i \,\}$ denote the set of anchors that share the same label of anchor $i$ (in our setting, anchors of the same instance). 
We then define the weighting function of the multi-positive term as follows:

{
\begin{equation}
\boldsymbol{w}_{i}
=2 - \dfrac{\exp\!\left(
\dfrac{1}{\sigma |\mathcal{B}_{y_i}|}\displaystyle\sum\limits_{j \in \mathcal{B}_{y_i}} \dfrac{1}{|P(i)|}\sum\limits_{r \in P(i)} s^{\scriptscriptstyle +}_{jr}\right)} 
{\displaystyle\sum\limits_{l \in \mathcal{B}}\dfrac{|P(l)|}{\sum\limits_{k \in \mathcal{B}} |P(k)|} 
\exp\!\left(
\dfrac{1}{\sigma |\mathcal{B}_{y_l}|}\sum\limits_{j \in \mathcal{B}_{y_l}} \dfrac{1}{|P(l)|}\sum\limits_{r \in P(l)} s^{\scriptscriptstyle +}_{jr}\right)} \;\;,
\end{equation}
}
where $\sigma$ is a temperature set to 0.5, and $\mathcal{B}$ denotes the batch. 
Subsequently, the \emph{decoupled weighted multi-positive contrastive loss} is:
\begin{equation}
\label{eq:decoupled_w_mu_pos_con_loss}
\mathcal{L}^{\textit{DWMP}}_{q_i}
=
-\dfrac{\boldsymbol{w}_{i}}{\tau|P(i)|}\sum\limits_{r\in P(i)}s^{\scriptscriptstyle +}_{ir} \;+\; \log\sum\limits_{j\in N(i)} \exp\!\left(s^{\scriptscriptstyle -}_{ij}/\tau\right) \;\;.
\end{equation}
We observe that, in the limit $\sigma \to \infty$, the weighting function satisfies $\boldsymbol{w}_{i}=1$, in which case the objective reduces to the \emph{decoupled multi-positive contrastive loss} $\mathcal{L}^{\textit{DMP}}_{q_i}$.

\noindent
{\bf Putting it all together.}
The $\ell_2$-normalized text, image and point-cloud features are co-aligned through the objective $\mathcal{L}_{Total}$ that is the weighted sum of three symmetric contrastive terms: 3D--Text, 3D--Image, and Text--Image:
\begin{equation}
\mathcal{L}_{Total}\;=\;\lambda_1\,\mathcal{L}_{3D\leftrightarrow T}\;+\;\lambda_2\,\mathcal{L}_{3D\leftrightarrow I}\;+\;\lambda_3\,\mathcal{L}_{T\leftrightarrow I},
\label{eq:Total_loss}
\end{equation}
where $\mathcal{L}_{A\leftrightarrow B}=\tfrac{1}{2}\big(\mathcal{L}_{A\to B}+\mathcal{L}_{B\to A}\big)$. Each directional term $\mathcal{L}_{A\to B}$ is a mean of per-anchor \emph{decoupled weighted multi-positive contrastive loss} of \cref{eq:decoupled_w_mu_pos_con_loss}, treating all matching views as positives and all in-batch non-matching views as negatives. 

\subsection{Text adapter}
\label{subsec:Text_source_alignment}

To cover the long tail and support open-vocabulary use, we require textual supervision that is both diverse and instance-grounded. 
Our data consists of three complementary text sources per object~\cite{NEURIPS2023_8c7304e7}, as seen in \cref{fig:input}: 
(i) annotations/metadata (concise, label-like names); 
(ii) VLM captions of rendered images, which describe the instance under a specific view; and 
(iii) captions of real web-images retrieved  from large image–text corpora using the rendered view as a query~\cite{schuhmann2022laion}, which inject broader, natural-language coverage of related real-world photos. 
In terms of fidelity, human annotations are most reliable. 
The captions of rendered images are moderately reliable, dependent on the specific view and the VLM used.
The retrieved web-captions are least reliable, as they describe semantically related but different real images.
Nevertheless, web-captions are valuable, offering broader natural text styles and finer-grained semantics than annotations or VLM captions.
Together, these sources supply varied phrasing, synonyms, and contextual cues that are especially valuable for rare categories.

Despite their benefits, these text sources come from different distributions.
Human annotations and captions of rendered images are highly correlated to the rendered images, as all describe the same instance, whereas web-captions describe similar but different real images and may be noisy or multilingual. 
This mismatch yields a domain gap (\eg, style and context), which impedes consistent cross-modal alignment. 
Consequently, the model underutilizes the web captions’ linguistic diversity and world knowledge, leaving a portion of their informative signal untapped during training.

To reconcile these distributions without discarding diversity, we introduce a lightweight MLP adapter applied only to retrieved web-caption embeddings. 
In practice, it is a two-layer MLP, fully connected layers with an intermediate nonlinearity and normalization followed by an output normalization and a residual skip connection back to the input to keep features stable.
The adapter is trained end-to-end with the objective to map web captions towards the rendered image embeddings, while leaving human annotations and captions of the rendered images untouched, as seen in \cref{fig:model}. 
This alignment reduces cross-domain drift and improves language to 3D distillation, yielding cleaner clusters and stronger zero-shot performance.

\section{Experiments}
\label{sec:experiments}
We evaluate the quality of our 3D point-cloud embeddings with two downstream tasks: zero-shot shape classification and cross-modal retrieval. 
Zero-shot shape classification provides a quantitative evaluation of separability and transferability, whereas cross-modal retrieval offers qualitative insight through example retrievals (see \cref{fig:teaser_image_3d_ret,fig:teaser_text_3d_ret}).
Together, these experiments validate that our embeddings are discriminative and cross-modal consistent, thereby supporting open-vocabulary recognition.
\cref{subsec:classification} presents zero-shot shape classification, while \cref{subsec:cross_modal_ret} presents the cross-modal retrieval application.

\noindent
\textbf{Pre-training datasets.}
We adopt the dataset from~\cite{NEURIPS2023_8c7304e7}, which constructed the input triplets, in the same manner as prior work~\cite{Zhang_2024_CVPR,Gao_2024_CVPR,Lei_2024_CVPR,qi2024shapellm, zhou2023uni3d}.
In this dataset, the training set consists of multi-modal tuples: a point cloud with multiple images and texts.
The images and texts were generated for four 3D datasets: ShapeNetCore \cite{chang2015shapenet}, 3D-FUTURE \cite{fu20213d}, ABO \cite{collins2022abo} and Objaverse~\cite{deitke2023objaverse}. 
Training was conducted on three subsets of the dataset:
(i) "ShapeNet", containing 52,470 tuples from ShapeNetCore; 
(ii) "Ensembled no LVIS", with 829,460 tuples drawn from all sources except the Objaverse-LVIS subset; 
and (iii) "Ensembled", comprising from all 875,665 available training tuples. 

\noindent
\textbf{Benchmark datasets.}
We evaluate our method on several benchmarks, following prior work: 
ModelNet40 \cite{wu20153d}, ScanObjectNN \cite{uy2019revisiting} and Objaverse-LVIS \cite{deitke2023objaverse}.
ModelNet40 is a CAD dataset covering 40 categories. 
ScanObjectNN consists of real-world point cloud objects extracted from indoor scenes across 15 categories. 
Objaverse-LVIS is a long-tail, annotated subset of Objaverse containing 46,832 shapes spaning across 1,156 categories derived from the LVIS dataset \cite{gupta2019lvis}.

\subsection{Zero-Shot Shape Classification}
\label{subsec:classification}

\begin{table*}[tb]
\caption{{\bf Zero-shot shape classification comparison.} 
HOLA outperforms other methods on most benchmarks under common pre-training settings and remains competitive on ModelNet40. 
The primary benchmark is Objaverse-LVIS, a large-scale dataset with a pronounced long-tail distribution, the core challenge our method targets. 
ScanObjectNN, a natural dataset whose distribution differs from that of the training data, provides an especially informative evaluation.
These gains are achieved with compact backbones, few parameters, and high computational efficiency (in terms of FLOPs).
\textbf{Bold} and \underline{underline} denote the best and second-best results, respectively.
}
\label{tab:zero_shot_comp_res}
\centering
\begin{adjustbox}{max width=\textwidth}
\setlength{\tabcolsep}{2pt}
\begin{tabular}{@{}l c | c | c | c c c | c c c | c c c@{}}
\toprule
\multicolumn{1}{c}{\multirow{2}{*}
{\textbf{Method}}} &
\multicolumn{1}{c|}{\multirow{2}{*}{\textbf{\#Params $\downarrow$}}} &
\multicolumn{1}{c|}{\multirow{2}{*}{\textbf{FLOPs $\downarrow$}}} &
\multicolumn{1}{c|}{\multirow{2}{*}{\textbf{FPS $\uparrow$}}} &
\multicolumn{3}{c|}{{\textbf{Objaverse-LVIS}}} &
\multicolumn{3}{c|}{\textbf{ScanObjectNN}} &
\multicolumn{3}{c}{\textbf{ModelNet40}} \\
& & & &
\textbf{Top-1} & \textbf{Top-3} & \textbf{Top-5} &
\textbf{Top-1} & \textbf{Top-3} & \textbf{Top-5} &
\textbf{Top-1} & \textbf{Top-3} & \textbf{Top-5} \\
\midrule
\multicolumn{13}{c}{\textit{Trained on "ShapeNet" (52,470 triplets)}} \\
\midrule
OpenShape-PointBERT \cite{NEURIPS2023_8c7304e7} & 5.1M & 1G & 240 & 10.8 & 20.2 & 25.0 & 51.3 & 69.4 & 78.4 & 70.3 & 86.9 & 91.3 \\
TAMM-PointBERT \cite{Zhang_2024_CVPR} & 35.4M & 29G & 85 & \underline{13.7} & \underline{24.2} & \underline{29.2} & \underline{54.8} & \underline{74.5} & \underline{83.3} & \underline{73.1} & \underline{88.5} & \underline{91.9} \\
\textbf{HOLA-PointBERT} & 32.3M & 29G & 202 & \textbf{17.7} & \textbf{30.0} & \textbf{36.1} & \textbf{55.9} & \textbf{77.4} & \textbf{87.7} & \textbf{74.2} & \textbf{90.0} & \textbf{95.0} \\
\midrule
\multicolumn{13}{c}{\textit{Trained on "Ensembled no LVIS" (829,460 triplets)}} \\
\midrule
OpenShape-PointBERT \cite{NEURIPS2023_8c7304e7} & 32.3M & 29G & 86 & 39.1 & 60.8 & 68.9 & 47.2 & 72.4 & 84.7 & 85.3 & 96.2 & 97.4 \\
TAMM-PointBERT \cite{Zhang_2024_CVPR} & 35.4M & 29G & 85 & 42.0 & 63.6 & 71.7 & 56.7 & 78.3 & 86.1 & 86.3 & 96.6 & 98.1 \\
VIT-LENS-G \cite{Lei_2024_CVPR} & 2000.0M & 1050G & 27 & 50.1 & 71.3 & \underline{78.1} & 59.8 & 79.3 & 87.7 & 86.8 & 96.8 & 97.8 \\
UNI3D-G \cite{zhou2023uni3d} & 1020.0M & 1130G & 34 & 47.2 & 68.8 & 76.1 & 66.5 & 83.5 & 90.1 & 86.8 & 97.3 & 98.4 \\
\textbf{HOLA-PointBERT} & 32.3M & 29G & 202 & \underline{50.3} & \underline{71.4} & 78.0 & \underline{67.2} & \textbf{85.4} & \underline{91.9} & \underline{88.3} & \textbf{97.6} & \underline{98.9} \\
\textbf{HOLA-PointBERT} & 72.1M & 84G & 152 & \textbf{50.7} & \textbf{72.1} & \textbf{78.9} & \textbf{67.7} & \textbf{85.4} & \textbf{92.4} & \textbf{89.0} & \textbf{97.6} & \textbf{99.0} \\
\midrule
\multicolumn{13}{c}{\textit{Trained on "Ensembled" (875,665 triplets)}} \\
\midrule
OpenShape-PointBERT \cite{NEURIPS2023_8c7304e7} & 32.3M & 29G & 86 & 46.8 & 69.1 & 77.0 & 52.2 & 79.7 & 88.7 & 84.4 & 96.5 & 98.0 \\
MixCon3D-PointBERT \cite{Gao_2024_CVPR} & 30.9M & 7G & 153 & 50.4 & 72.2 & 79.1 & 58.6 & 80.3 & 89.2 & 86.8 & 96.9 & 98.3 \\
TAMM-PointBERT \cite{Zhang_2024_CVPR} & 35.4M & 29G & 85 & 50.7 & 73.2 & 80.6 & 55.7 & 80.7 & 88.9 & 85.0 & 96.6 & 98.1 \\
VIT-LENS-G \cite{Lei_2024_CVPR} & 1543.0M & 806G & 32 & 52.0 & 73.3 & 79.9 & 60.1 & 81.0 & 90.3 & 87.6 & 96.6 & 98.4 \\
UNI3D-G \cite{zhou2023uni3d} & 1020.0M & 1130G & 34 & 55.3 & 76.7 & 82.9 & 65.3 & 85.5 & 92.7 & \underline{88.2} & \textbf{98.4} & \textbf{99.3} \\
RECON++-B \cite{qi2024shapellm} & 201.5M & 137G & 90 & 53.2 & 75.3 & 81.5 & 63.6 & 80.2 & 90.6 & 86.5 & 94.7 & 95.8 \\
RECON++-L \cite{qi2024shapellm} & 658.9M & 423G & 34 & 53.7 & 75.8 & 82.0 & 65.4 & 84.1 & 89.7 & 87.3 & 95.4 & 96.1 \\
\textbf{HOLA-PointBERT} & 26.0M & 7G & 264 & 55.7 & 77.3 & 83.6 & 65.1 & 83.7 & 90.9 & 87.5 & 97.4 & 98.7 \\
\textbf{HOLA-PointBERT} & 32.3M & 29G & 202 & \underline{56.7} & \underline{78.4} & \underline{84.4} & \underline{68.3} & \textbf{86.7} & \underline{93.0} & 88.1 & 97.4 & 98.6 \\
\textbf{HOLA-PointBERT} & 72.1M & 84G & 152 & \textbf{57.3} & \textbf{79.0} & \textbf{84.9} & \textbf{68.7} & \underline{86.5} & \textbf{93.1} & \textbf{88.8} & \underline{97.6} & \underline{98.8} \\
\bottomrule
\end{tabular}
\end{adjustbox}
\end{table*}

Zero-shot shape classification aims to categorize 3D shapes into semantic classes not seen during training. 
The model learns a shared cross-modal representation space, allowing it to generalize to unseen categories at test time.
This setting is crucial for long-tail distributions, where many classes are rare or absent during training. 
It enables recognition of novel shapes without labeled examples, reducing annotation costs and improving generalization to real-world open-set scenarios.

Current work can be roughly categorized into two groups: approaches built on compact 3D backbones (\eg, PointBERT \cite{yu2022point} and MinkowskiNet \cite{choy20194d}) and those that utilize large 2D ViTs \cite{dosovitskiy2021an}, often with one or two orders of magnitude more parameters. 
The latter methods consistently outperform alternatives by a significant margin.
Our method is the first to reverse this trend, achieving stronger performance than ViT-based methods while relying only on compact 3D backbones.

\Cref{tab:zero_shot_comp_res} reports our results with the PointBERT 3D backbone on the standard benchmarks: Objaverse-LVIS \cite{deitke2023objaverse}, ScanObjectNN (OBJ\_ONLY) \cite{uy2019revisiting}, and ModelNet40 \cite{wu20153d}; corresponding MinkowskiNet results are provided in \cref{supp_sec:additional_exp} of the supplementary material.  
Among these benchmarks, Objaverse-LVIS is especially important, as its pronounced long-tail distribution provides the most informative measure of open-world robustness.
On this challenging benchmark, using the “Ensembled” training set, our 72M-parameter model surpasses the previous best method \cite{zhou2023uni3d} by $2.0\%$, despite its reliance on a much larger 1020M-parameter model; even our smaller 26M-parameter variant still outperforms it by $0.4\%$.
We also outperform the strongest method with a small 3D backbone \cite{Zhang_2024_CVPR} by $6.0\%$ using a comparable model size (32M), indicating that these gains arise primarily from our loss design rather than increased model capacity.
Moreover, using the same training set, our models exhibit strong transferability: on ScanObjectNN, a real-scan dataset with a pronounced domain gap from our training data, our 72M-parameter model improves over the best reported result \cite{qi2024shapellm} by $3.3\%$, while our 26M-parameter model outperforms the strongest method with a small 3D backbone \cite{Gao_2024_CVPR} by $6.5\%$. 

Another direct merit of our approach, enabled by the lightweight model design, is its computational efficiency. 
As shown in \Cref{tab:zero_shot_comp_res}, our method is the first to surpass the best ViT-based approaches while using lightweight 3D backbones with fewer than 72M parameters, substantially smaller than the 200M-2000M parameters of competing models.

Beyond network size, our models require substantially fewer FLOPs and achieve the lowest inference latency, resulting in the highest throughput (FPS) among all compared methods. Overall, they are compute- and latency-efficient.

\begin{figure*}[t]
    \centering
    \includegraphics[width=\linewidth, height=0.31\textheight]{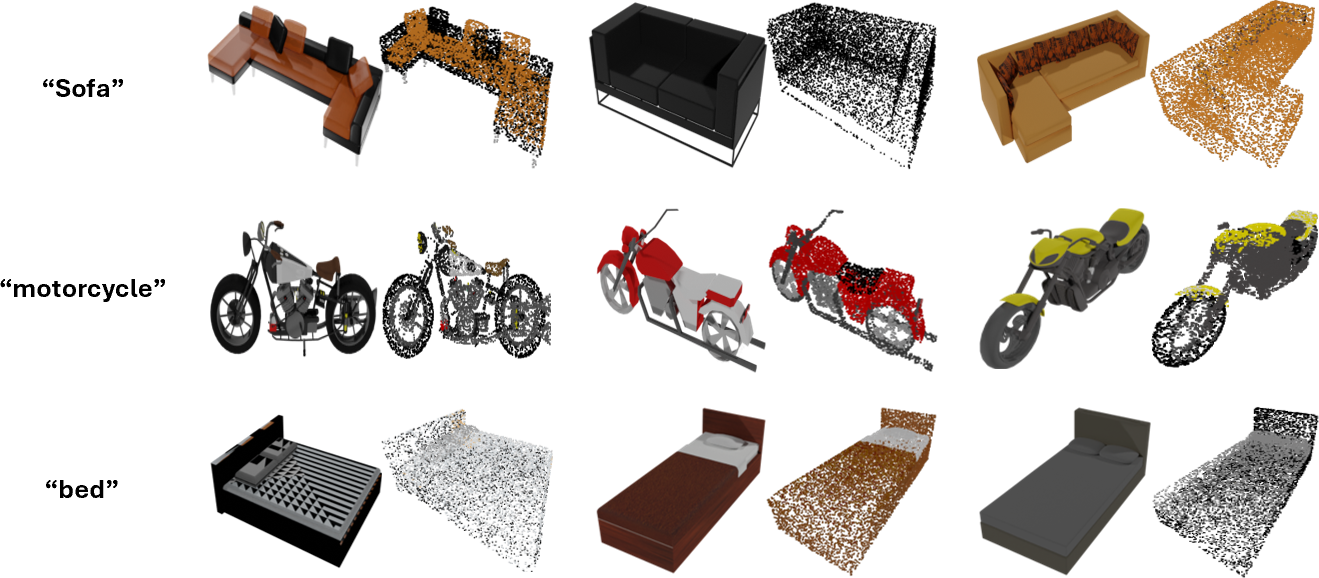} \\
    \hspace{0.05in} Input \hspace{0.475in} 1st Retrieved Result \hspace{0.4in} 2nd Retrieved Result \hspace{0.375in} 3rd Retrieved Result
    \caption{
    \textbf{Text-to-3D retrieval for common (head) categories.}
    Each retrieved result is shown as a pair: an RGB rendering of the retrieved shape and its corresponding point cloud, which is the item returned by the search.
    The results indicate a well-structured embedding space across head categories.
    }
    \label{fig:ret_txt_head_cat}
\end{figure*}

\subsection{Cross-Modal Retrieval}
\label{subsec:cross_modal_ret}
Cross-modal retrieval \cite{li2015joint,chen2018text2shape,wang2025cross} evaluates how faithfully our shared embedding space aligns modalities.
It probes whether the shared embedding space enables practical search, \ie, given a text or image query, can the system retrieve the correct 3D shape without task-specific fine-tuning.
At evaluation, we embed the query with the frozen image/text encoder and embed all candidate point clouds with the 3D encoder. 
After \(\ell_2\) normalization, we compute cosine similarity and return the k-nearest 3D neighbors (text-to-3D and image-to-3D). 
Following standard practice, we conduct cross-modal retrieval experiments on the “Ensembled” set.

\begin{figure*}[t]
    \centering
    \includegraphics[width=\linewidth,height=0.44\textheight]{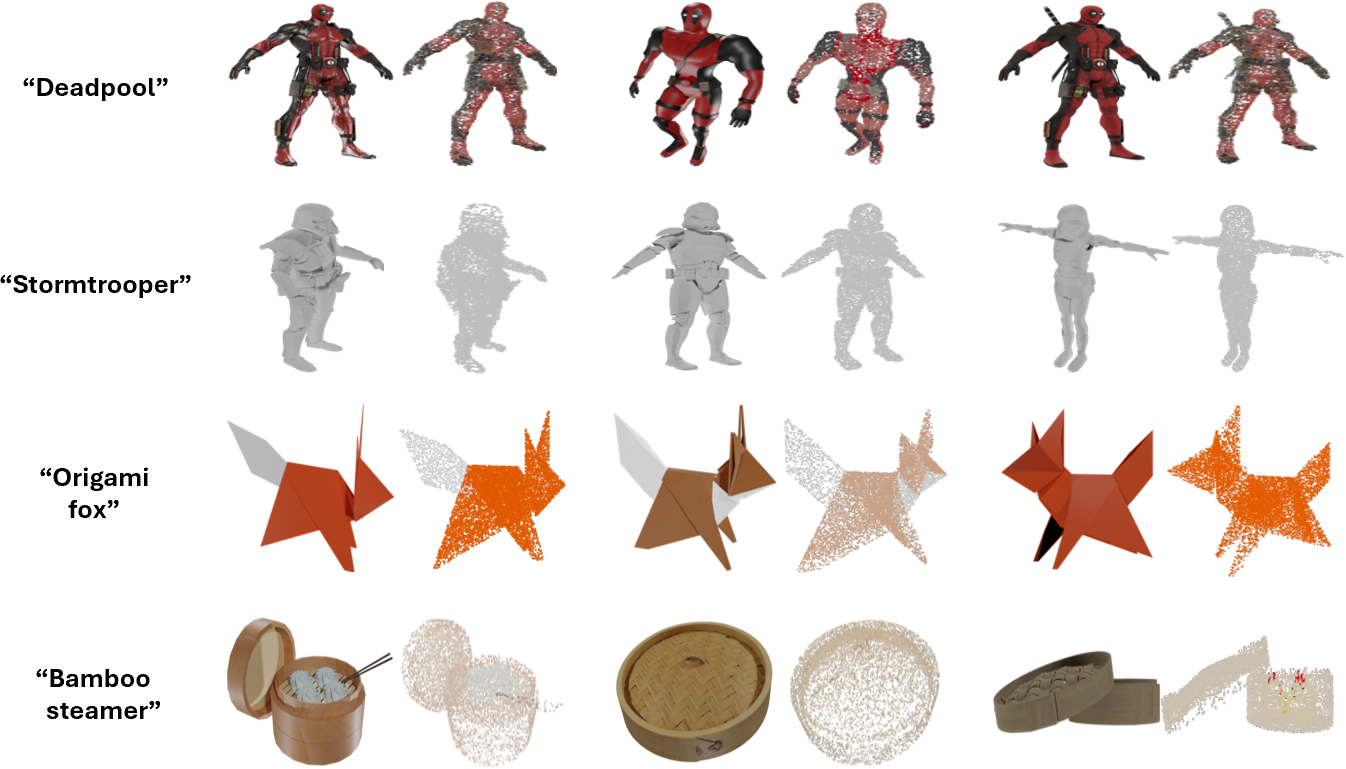} \\
    \hspace{0.05in} Input \hspace{0.475in} 1st Retrieved Result \hspace{0.35in} 2nd Retrieved Result \hspace{0.35in} 3rd Retrieved Result
    \caption{
    \textbf{Text-to-3D retrieval for rare (tail) categories.}
    Each retrieved result is shown as a pair: an RGB rendering of the retrieved shape and its corresponding point cloud, which is the item returned by the search.
    The results indicate a well-structured embedding space with robust performance on long-tail classes.
    }
    \label{fig:ret_txt_tail_cat}
\end{figure*}

As illustrated in \cref{fig:teaser_text_3d_ret,fig:ret_txt_head_cat,fig:ret_txt_tail_cat}, our method enables successful text-to-3D retrieval for both common (head) and rare (tail) categories. 
These qualitative results reinforce the quantitative gains on the long-tailed Objaverse-LVIS benchmark reported in \Cref{tab:zero_shot_comp_res}. 
They further support the effectiveness of our approach for open-vocabulary generalization, particularly on unseen and long-tail classes. 
At the same time, they show that our method preserves strong performance on common categories.
Additional text- and image-based 3D retrieval results are provided in \Cref{supp_sec:additional_exp} of the supplementary.

\subsection{Ablation study}
\label{subsec:ablation}
In this section, we evaluate the effectiveness of our decoupled multi-positive contrastive loss, the weighting function for the multi-positive alignment term, the text adapter, and the effects of multiple text and image views.
Throughout this section, we report results using a 32M-parameter PointBERT backbone trained on the "ShapeNet" training set, under the default setting of eight views for both the image and text modalities, unless stated otherwise.
Further ablations, including temperature, batch size, model scaling, and additional design choices, are provided in \cref{supp_sec:additional_ablations} of the supplementary material.

\noindent
\textbf{Decoupled multi-positive loss.}
\Cref{tab:zero_shot_comp_mp_vs_dmp} shows that replacing the naive multi-positive (or single-positive) loss with our decoupled formulation improves Objaverse-LVIS Top-1 accuracy. 
In particular, with eight views, the gain reaches $3.1\%$ on "ShapeNet" and $5.4\%$ on "Ensembled no LVIS".

Furthermore, under the naive multi-positive loss, increasing the number of views unexpectedly degrades performance: the single-view setting performs best, and accuracy declines as more views are added. 
This behavior can be attributed to the “spotlight crowding” phenomenon (see \cref{fig:spotlight_crowding}). 
Our proposed loss resolves this issue, and performance improves consistently as additional views are added.

\begin{table*}[tb]
\caption{
{\bf MP vs. DMP.}
Our decoupled multi-positive loss consistently outperforms the multi-positive baseline at every view count, with gains increasing as more views are added. 
Values in parentheses indicate the improvement margin at each view count on Objaverse-LVIS Top-1 zero-shot accuracy. 
Experiments use a 32M-parameter model without the text adapter.
}
\label{tab:zero_shot_comp_mp_vs_dmp}
\centering
\begin{adjustbox}{max width=\textwidth}
\setlength{\tabcolsep}{2.5pt}
\begin{tabular}{@{}c c c c c | c c c c@{}}
\toprule
\multicolumn{1}{c}{\multirow{2}{*}{\textbf{Dataset}}} &
\multicolumn{4}{c|}
{\textbf{Multi-Positive}} &
\multicolumn{4}{c}
{\textbf{Decoupled Multi-Positive}} \\
 & 1-view & 4-view & 8-view & 12-view & 1-view & 4-view & 8-view & 12-view \\
\midrule
\textbf{ShapeNet} & 13.8 & 13.6 & 13.4 & 13.3 & 15.8 ({\bf \color{forestgreen}{+2.0\%}}) & 16.2 ({\bf \color{forestgreen}{+2.6\%}}) & 16.5 ({\bf \color{forestgreen}{+3.1\%}}) & 16.4 ({\bf \color{forestgreen}{+3.1\%}}) \\
\textbf{Ensembled no LVIS} & 45.1 & 44.9 & 44.4 & 44.6 & 49.4 ({\bf \color{forestgreen}{+4.3\%}}) & 49.6 ({\bf \color{forestgreen}{+4.7\%}}) & 49.8 ({\bf \color{forestgreen}{+5.4\%}}) & 49.8 ({\bf \color{forestgreen}{+5.2\%}}) \\
\bottomrule
\end{tabular}
\end{adjustbox}
\end{table*}

\begin{table}[tb]
\caption{{\bf Effect of the weighting function.} Incorporating weighting function in the multi-positive alignment term yields consistent accuracy gains across benchmarks.}
\label{tab:weightning_func}
\centering
\small
\begin{adjustbox}{max width=\columnwidth}
\begin{tabular}{@{}c | c c c | c c c | c c c@{}}
\toprule
\multicolumn{1}{c|}{\multirow{2}{*}{\shortstack[c]{\textbf{Weighting}\\\textbf{Function}}}} &
\multicolumn{3}{c|}{\textbf{Objaverse-LVIS}} &
\multicolumn{3}{c|}{\textbf{ScanObjectNN}} &
\multicolumn{3}{c}{\textbf{ModelNet40}} \\
& 
\textbf{Top-1} & \textbf{Top-3} & \textbf{Top-5} &
\textbf{Top-1} & \textbf{Top-3} & \textbf{Top-5} &
\textbf{Top-1} & \textbf{Top-3} & \textbf{Top-5}\\
\midrule
\xmark & 16.9 & 29.0 & 35.1 & 55.4 & 76.6 & 86.7 & 73.6 & \textbf{90.8} & 94.7 \\
\cmark & \textbf{17.7} & \textbf{30.0} & \textbf{36.1} & \textbf{55.9} & \textbf{77.4} & \textbf{87.7} & \textbf{74.2} & 90.0 & \textbf{95.0} \\
\bottomrule
\end{tabular}
\end{adjustbox}
\end{table}

\noindent
\textbf{Weighting function for the multi-positive alignment term.}
As \Cref{tab:weightning_func} indicates, adding a weighting function to the multi-positive term improves Top-1 accuracy across all benchmarks. 
This is expected, as the weighting function makes the positive term hardness-aware, emphasizing anchors that remain farther from their positives and thus require stronger corrective updates, while down-weighting already well-aligned anchors. 
As a result, the optimization focuses on the most informative cases, leading to more effective multi-positive alignment and consistently better performance.

\begin{table}[t]
\caption{{\bf Effect of the text adapter.} 
Incorporating the text adapter improves accuracy across all benchmarks and metrics.
}
\label{tab:text_adapter}
\centering
\small
\begin{adjustbox}{max width=\columnwidth}
\begin{tabular}{@{}c | c c c | c c c | c c c@{}}
\toprule
\multicolumn{1}{c|}{\multirow{2}{*}{\shortstack[c]{\textbf{Text}\\\textbf{Adapter}}}} &
\multicolumn{3}{c|}{{\textbf{Objaverse-LVIS}}} &
\multicolumn{3}{c|}{\textbf{ScanObjectNN}} &
\multicolumn{3}{c}{\textbf{ModelNet40}} \\
&
\textbf{Top-1} & \textbf{Top-3} & \textbf{Top-5} &
\textbf{Top-1} & \textbf{Top-3} & \textbf{Top-5} &
\textbf{Top-1} & \textbf{Top-3} & \textbf{Top-5}\\
\midrule
\xmark & 17.1 & 29.5 & 35.5 & 51.9 & 74.7 & 86.6 & 73.5 & 88.0 & 92.3 \\
\cmark & \textbf{17.7} & \textbf{30.0} & \textbf{36.1} & \textbf{55.9} & \textbf{77.4} & \textbf{87.7} & \textbf{74.2} & \textbf{90.0} & \textbf{95.0} \\
\bottomrule
\end{tabular}
\end{adjustbox}
\end{table}

\noindent
\textbf{Text adapter.}
\Cref{tab:text_adapter} shows that incorporating a lightweight text adapter, applied only to retrieved web captions, consistently boosts performance, improving accuracy across all datasets and metrics. For example, it increases Top-1 accuracy by $0.6\%$ on the  Objaverse-LVIS dataset and by $4.0\%$ on real scanned objects from ScanObjectNN.

This occurs because the text adapter narrows the domain gap between the retrieved web-caption embeddings and the rendered image embeddings. Consequently, it enhances the alignment of web-caption embeddings with both annotation and VLM-generated caption embeddings, thereby improving the overall 3D–language alignment.

\noindent
\textbf{Multiple textual and image views.}
\Cref{tab:optimal_view_selection} shows that incorporating complementary views improves zero-shot performance. 
Using eight views yields the best results, increasing Top-1 accuracy on the challenging Objaverse-LVIS benchmark by $0.9\%$ over the standard single-view setting.
This gain stems from the fact that any single rendering or caption captures only a partial, view-biased slice of a 3D object, potentially omitting self-occluded regions, fine geometry, or texture cues. 
Aggregating images and texts from diverse viewpoints provides complementary evidence, enforces cross-view consistency, and anchors the representation to the holistic shape rather than to a single view.

Furthermore, \Cref{tab:multiple_text_views,tab:multiple_image_views} show that both text and image modalities benefit from multiple views.
With image views fixed at eight, increasing text views steadily improves Top-1 zero-shot accuracy on the challenging Objaverse-LVIS benchmark. 
Moving from one to four views yields a $0.8\%$ gain, and extending to eight views adds another $0.4\%$.
A similar trend holds for image views: with text views fixed at eight, increasing image views from one to four improves Top-1 zero-shot accuracy on Objaverse-LVIS by $0.6\%$, and extending to eight views yields a further $0.2\%$ gain.
These gains show that text and image views contribute complementary, non-redundant cues. Aggregating multiple views in both modalities reduces single-view bias and strengthens cross-modal alignment, improving zero-shot accuracy on the Objaverse-LVIS long-tail benchmark.

\begin{table}[tb]
\caption{{\bf Effect of the number of views.} 
Zero-shot accuracy improves with additional views, reaching its best at eight views.
}
\label{tab:optimal_view_selection}
\centering
\small
\begin{adjustbox}{max width=\columnwidth}
\begin{tabular}{@{}c | c c c | c c c | c c c@{}}
\toprule
\multicolumn{1}{c|}{\multirow{2}{*}{\textbf{\#Views}}} &
\multicolumn{3}{c|}{\textbf{Objaverse-LVIS}} &
\multicolumn{3}{c|}{\textbf{ScanObjectNN}} &
\multicolumn{3}{c}{\textbf{ModelNet40}} \\
& 
\textbf{Top-1} & \textbf{Top-3} & \textbf{Top-5} &
\textbf{Top-1} & \textbf{Top-3} & \textbf{Top-5} &
\textbf{Top-1} & \textbf{Top-3} & \textbf{Top-5}\\
\midrule
1 & 16.8 & 28.7 & 34.5 & 54.6 & 75.0 & 84.5 & 73.7 & 89.4 & 94.2 \\
2 & 16.8 & 28.6 & 34.6 & 54.7 & 77.1 & 86.7 & \textbf{74.3} & 89.2 & 93.2 \\
4 & 16.9 & 29.2 & 35.2 & 55.1 & 76.6 & 86.4 & 73.3 & 89.8 & 94.4 \\
6 & 17.2 & 29.4 & 35.5 & 53.1 & 74.9 & 85.9 & 73.7 & 89.4 & 93.8 \\
8 & \textbf{17.7} & \textbf{30.0} & \textbf{36.1} & \textbf{55.9} & \textbf{77.4} & \textbf{87.7} & 74.2 & 90.0 & \textbf{95.0} \\
10 & \textbf{17.7} & 29.9 & 35.8 & 54.1 & 75.7 & 85.5 & 73.1 & 90.4 & \textbf{95.0} \\
12 & 17.3 & 29.8 & 35.8 & 55.1 & 77.2 & 86.8 & 74.0 & \textbf{90.9} & 94.8 \\
\bottomrule
\end{tabular}
\end{adjustbox}
\end{table}

\begin{table}[tb]
\caption{{\bf Effect of using multiple texts.} On the challenging Objaverse-LVIS benchmark, increasing the number of text views while holding the number of image views fixed, improves accuracy.}
\label{tab:multiple_text_views}
\centering
\small
\begin{adjustbox}{max width=\columnwidth}
\begin{tabular}{@{}c c | c c c | c c c | c c c@{}}
\toprule
\multicolumn{2}{c|}{\textbf{\#Views}} &
\multicolumn{3}{c|}{\textbf{Objaverse-LVIS}} &
\multicolumn{3}{c|}{\textbf{ScanObjectNN}} &
\multicolumn{3}{c}{\textbf{ModelNet40}} \\
\textbf{Text} & \textbf{Image} &
\textbf{Top-1} & \textbf{Top-3} & \textbf{Top-5} &
\textbf{Top-1} & \textbf{Top-3} & \textbf{Top-5} &
\textbf{Top-1} & \textbf{Top-3} & \textbf{Top-5}\\
\midrule
1 & 8 & 16.5 & 28.5 & 34.1 & 53.0 & 76.6 & 85.8 & 73.3 & \textbf{90.6} & \textbf{95.6} \\
4 & 8 & 17.3 & 29.2 & 35.0 & 52.8 & 74.6 & 85.2 & 73.5 & 87.1 & 91.3 \\
8 & 8 & \textbf{17.7} & \textbf{30.0} & \textbf{36.1} & \textbf{55.9} & \textbf{77.4} & \textbf{87.7} & \textbf{74.2} & 90.0 & 95.0 \\
\bottomrule
\end{tabular}
\end{adjustbox}
\end{table}

\begin{table}[tb]
\caption{{\bf Effect of using multiple images.} 
Increasing the number of image views while holding the number of text views fixed, improves accuracy.}
\label{tab:multiple_image_views}
\centering
\small
\begin{adjustbox}{max width=\columnwidth}
\begin{tabular}{@{}c c | c c c | c c c | c c c@{}}
\toprule
\multicolumn{2}{c|}{\textbf{\#Views}} &
\multicolumn{3}{c|}{\textbf{Objaverse-LVIS}} &
\multicolumn{3}{c|}{\textbf{ScanObjectNN}} &
\multicolumn{3}{c}{\textbf{ModelNet40}} \\
\textbf{Text} & \textbf{Image} &
\textbf{Top-1} & \textbf{Top-3} & \textbf{Top-5} &
\textbf{Top-1} & \textbf{Top-3} & \textbf{Top-5} &
\textbf{Top-1} & \textbf{Top-3} & \textbf{Top-5}\\
\midrule
8 & 1 & 16.9 & 29.0 & 35.1 & 55.1 & 76.9 & 86.3 & 73.6 & 89.6 & 94.4 \\
8 & 4 & 17.5 & 29.7 & 35.7 & 54.5 & 75.8 & 86.5 & 73.8 & 88.9 & 94.2 \\
8 & 8 & \textbf{17.7} & \textbf{30.0} & \textbf{36.1} & \textbf{55.9} & \textbf{77.4} & \textbf{87.7} & \textbf{74.2} & \textbf{90.0} & \textbf{95.0} \\
\bottomrule
\end{tabular}
\end{adjustbox}
\end{table}

\noindent
\textbf{Limitations.}
Our models' performance on ModelNet40 (Top-3 and Top-5) is slightly below SoTA (see \Cref{tab:zero_shot_comp_res}).
This can be attributed to the nature of ModelNet40: a small and saturated CAD dataset lacking natural categories and long-tail distributions.
Nevertheless, our method attains comparable Top-1 accuracy, demonstrating strong generalization ability.

\section{Conclusion}
This paper presents a method for open-set 3D recognition that handles rare and unseen categories.
The method aligns each point cloud with multiple images and textual descriptions to obtain a holistic representation.
A key contribution is the proposed {\em decoupled multi-positive contrastive loss,} which jointly aligns a 3D instance with its matched multi-view images and multiple texts, while decoupling positive aggregation from negative competition. 
This formulation preserves the loss’s hardness-aware focus on challenging negatives and mitigates the “spotlight crowding” that arises when many positives share the same softmax with all negatives. 
In addition, the method incorporates a lightweight text adapter applied only to web captions, thereby reducing the domain gap to curated annotations.

Our models achieve state-of-the-art open-vocabulary performance on long-tail benchmarks, improving zero-shot accuracy while maintaining high frame rates.
This combination of accuracy and efficiency makes them well suited to real-world deployment under tight computational and memory constraints, bridging the gap between first-in-class methods and truly deployable systems.

{
    \small
    \bibliographystyle{plainnat}
    \bibliography{references}
}

\clearpage
\appendix
\renewcommand{\theHsection}{appendix.\Alph{section}}

\begin{center}
    {\LARGE \bfseries Supplementary Material \par}
    \vspace{2em}
\end{center}

\noindent
\textbf{Supplementary material overview.}
\Cref{supp_sec:additional_exp} extends the zero-shot comparisons to an additional 3D backbone and provides additional results for the cross-modal retrieval application.
\Cref{supp_sec:analysis_of_hardness_aware_simple_loss} analyzes how different hardness-aware weighting regimes affect performance in a simple contrastive learning setting.
\Cref{supp_sec:additional_ablations} reports additional ablations on the key components of our method.
\Cref{supp_sec:implementation_details} provides implementation details and hyper-parameters selection. 

\begin{table*}[b]
\caption{{\bf Zero-shot shape classification comparison for SparseConv \cite{choy20194d}.} 
HOLA outperforms competing methods across most benchmarks and standard pretraining regimes, with only a few metrics on ScanObjectNN and ModelNet40 where it remains competitive. 
Our primary focus is Objaverse-LVIS, a large, long-tail benchmark that best reflects the core challenge our approach is designed to address.
}
\label{tab:zero_shot_comp_res_sparse_conv}
\centering
\begin{adjustbox}{max width=\textwidth}
\setlength{\tabcolsep}{3.5pt}
\begin{tabular}{@{}l c c c | c c c | c c c@{}}
\toprule
\multicolumn{1}{c}{\multirow{2}{*}{\textbf{Method}}} &
\multicolumn{3}{c|}{{\textbf{Objaverse-LVIS}}} &
\multicolumn{3}{c|}{\textbf{ScanObjectNN}} &
\multicolumn{3}{c}{\textbf{ModelNet40}} \\
&
\textbf{Top-1} & \textbf{Top-3} & \textbf{Top-5} &
\textbf{Top-1} & \textbf{Top-3} & \textbf{Top-5} &
\textbf{Top-1} & \textbf{Top-3} & \textbf{Top-5} \\
\midrule
\multicolumn{10}{c}{\textit{Trained on "ShapeNet" (52,470 triplets)}} \\
\midrule
OpenShape-SparseConv \cite{NEURIPS2023_8c7304e7} & 11.6 & 21.8 & 27.1 & 52.7 & 72.7 & 83.6 & 72.9 & 87.2 & 93.0 \\
TAMM-SparseConv \cite{Zhang_2024_CVPR} & \underline{13.6} & \underline{24.2} & \underline{29.3} & \textbf{57.9} & \underline{75.3} & \underline{83.1} & \underline{74.6} & \underline{88.2} & \underline{94.0} \\
\textbf{HOLA-SparseConv} & \textbf{17.5} & \textbf{30.0} & \textbf{36.1} & \underline{56.3} & \textbf{77.4} & \textbf{87.6} & \textbf{75.4} & \textbf{90.2} & \textbf{94.8} \\
\midrule
\multicolumn{10}{c}{\textit{Trained on "Ensembled no LVIS" (829,460 triplets)}} \\
\midrule
OpenShape-SparseConv \cite{NEURIPS2023_8c7304e7} & 37.0 & 58.4 & 66.9 & 54.9 & 76.8 & 87.0 & 82.6 & 95.0 & 97.5 \\
TAMM-SparseConv \cite{Zhang_2024_CVPR} & \underline{39.8} & \underline{62.0} & \underline{70.4} & \underline{57.5} & \textbf{81.3} & \textbf{90.0} & \textbf{85.7} & \textbf{96.8} & \textbf{98.3} \\
\textbf{HOLA-SparseConv} & \textbf{42.9} & \textbf{65.2} & \textbf{72.9} & \textbf{59.3} & \underline{79.8} & \underline{89.4} & \underline{85.3} & \underline{96.2} & \underline{98.1} \\
\midrule
\multicolumn{10}{c}{\textit{Trained on "Ensembled" (875,665 triplets)}} \\
\midrule
OpenShape-SparseConv \cite{NEURIPS2023_8c7304e7} & 43.4 & 64.8 & 72.4 & 56.7 & 78.9 & 88.6 & 83.4 & 95.6 & 97.8 \\
TAMM-SparseConv \cite{Zhang_2024_CVPR} & \underline{43.8} & \underline{66.2} & \underline{74.1} & \underline{58.5} & \underline{81.3} & \underline{89.5} & \underline{85.4} & \textbf{96.4} & \textbf{98.1} \\
\textbf{HOLA-SparseConv} & \textbf{47.2} & \textbf{69.3} & \textbf{76.3} & \textbf{60.6} & \textbf{81.4} & \textbf{91.0} & \textbf{85.5} & \underline{95.9} & \underline{98.0} \\
\bottomrule
\end{tabular}
\end{adjustbox}
\end{table*}

\section{Additional Experiments}
\label{supp_sec:additional_exp}

\noindent
\textbf{Zero-shot shape classification for SparseConv \cite{choy20194d}.}
\Cref{tab:zero_shot_comp_res_sparse_conv} compares methods that use a sparse-convolutional 3D backbone (\ie, MinkowskiNet), complementing the PointBERT \cite{yu2022point} results in \Cref{tab:zero_shot_comp_res}. 
All methods employ the same SparseConv variant. 
For the "Ensembled" and "Ensembled no LVIS" settings, our models were trained on 8 GPUs with a per-GPU batch size of 64 (global batch size 512).

As shown in \Cref{tab:zero_shot_comp_res_sparse_conv}, on the challenging long-tail Objaverse-LVIS benchmark, our method attains the strongest zero-shot results, exceeding competitors by at least $3.1\%$ in top-1 accuracy across all configurations.
Such substantial gains further indicate that our method is especially effective at handling rare and unseen categories. 
Moreover, under the “Ensembled” and “Ensembled no LVIS” pretraining regimes, our method achieves at least a $1.8\%$ improvement in Top-1 accuracy over competing methods on the real-scan ScanObjectNN dataset.

\begin{figure*}[tb]
    \centering
    \includegraphics[width=\linewidth,height=0.32\textheight]{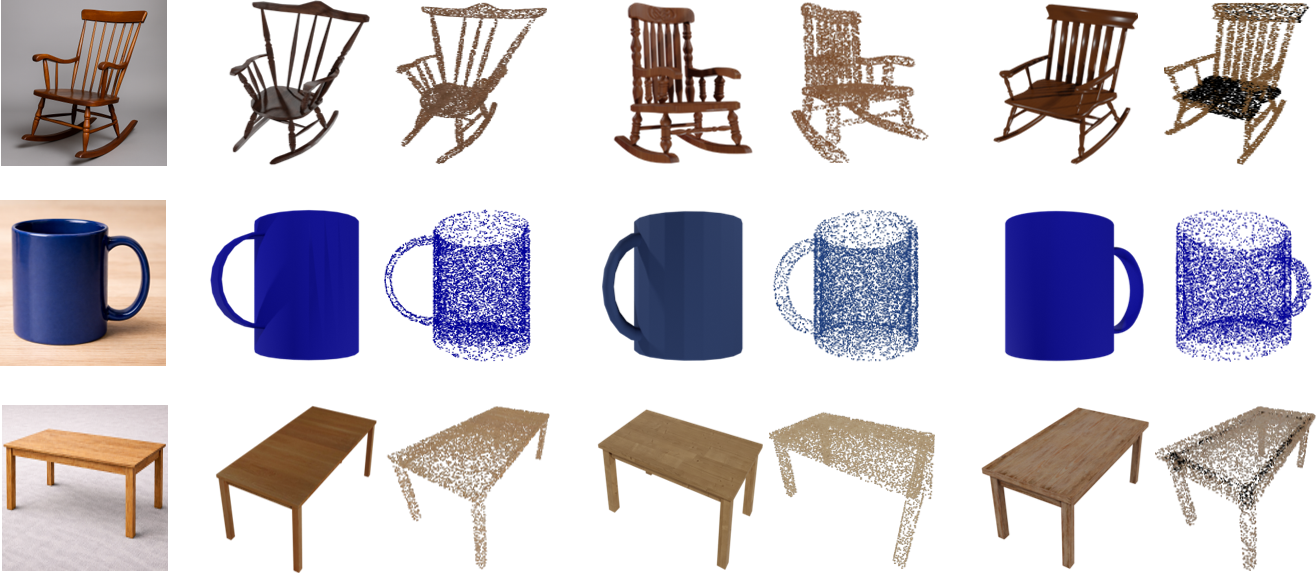} \\
    \hspace{0.075in} Input \hspace{0.475in} 1st Retrieved Result \hspace{0.375in} 2nd Retrieved Result \hspace{0.4in} 3rd Retrieved Result
    \caption{
    \textbf{Image-to-3D retrieval for common (head) categories.}
    Each retrieved result is shown as a pair: an RGB rendering of the retrieved shape and its corresponding point cloud, which is the item returned by the search.
    The results indicate a well-structured embedding space across head categories.
    }
    \label{fig:ret_img_head_cat}
\end{figure*}

\begin{figure*}[tb]
    \centering
    \includegraphics[width=\linewidth,height=0.44\textheight]{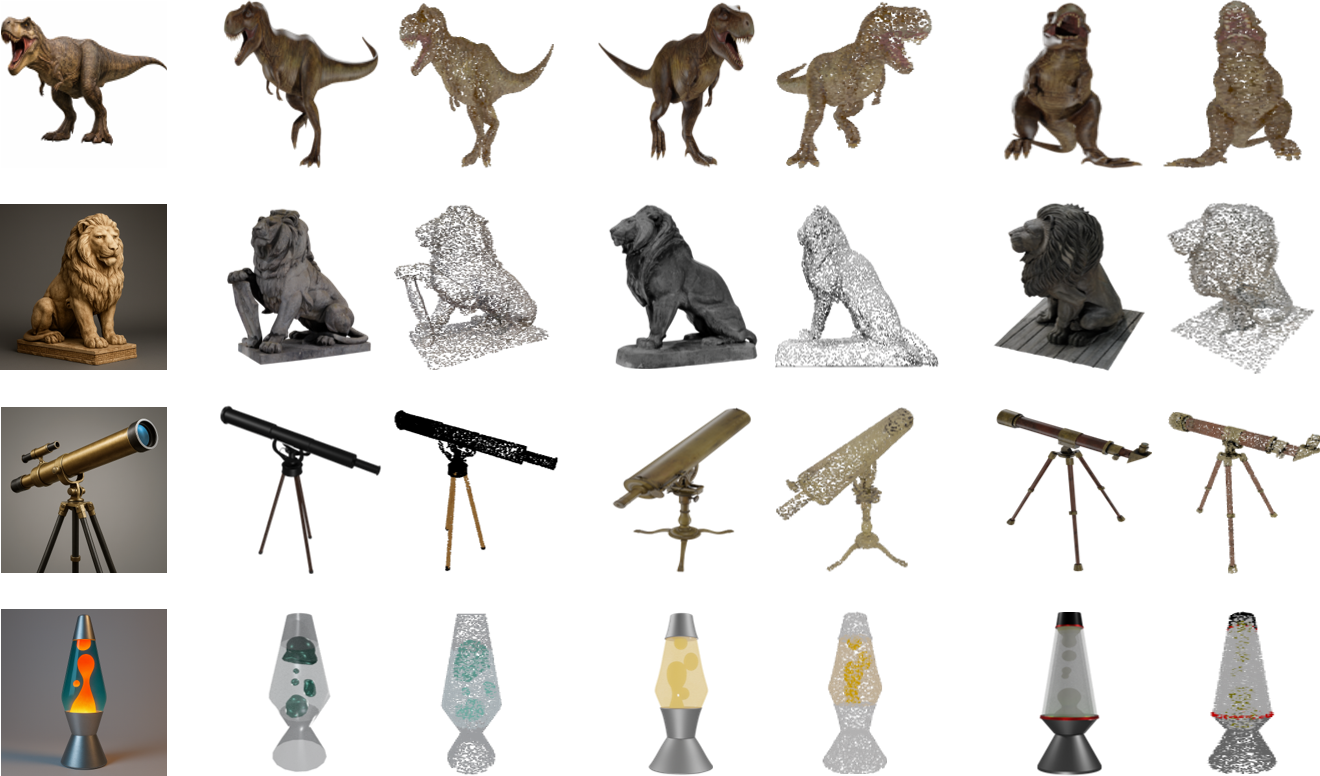} \\
    \hspace{0.045in} Input \hspace{0.5in} 1st Retrieved Result \hspace{0.36in} 2nd Retrieved Result \hspace{0.4in} 3rd Retrieved Result
    \caption{
    \textbf{Image-to-3D retrieval for rare (tail) categories.}
    Each retrieved result is shown as a pair: an RGB rendering of the retrieved shape and its corresponding point cloud, which is the item returned by the search.
    The results indicate a well-structured embedding space with robust performance on long-tail classes.
    }
    \label{fig:ret_img_tail_cat}
\end{figure*}

\begin{figure*}[tb]
    \centering
    \includegraphics[width=\linewidth,height=0.39\textheight]{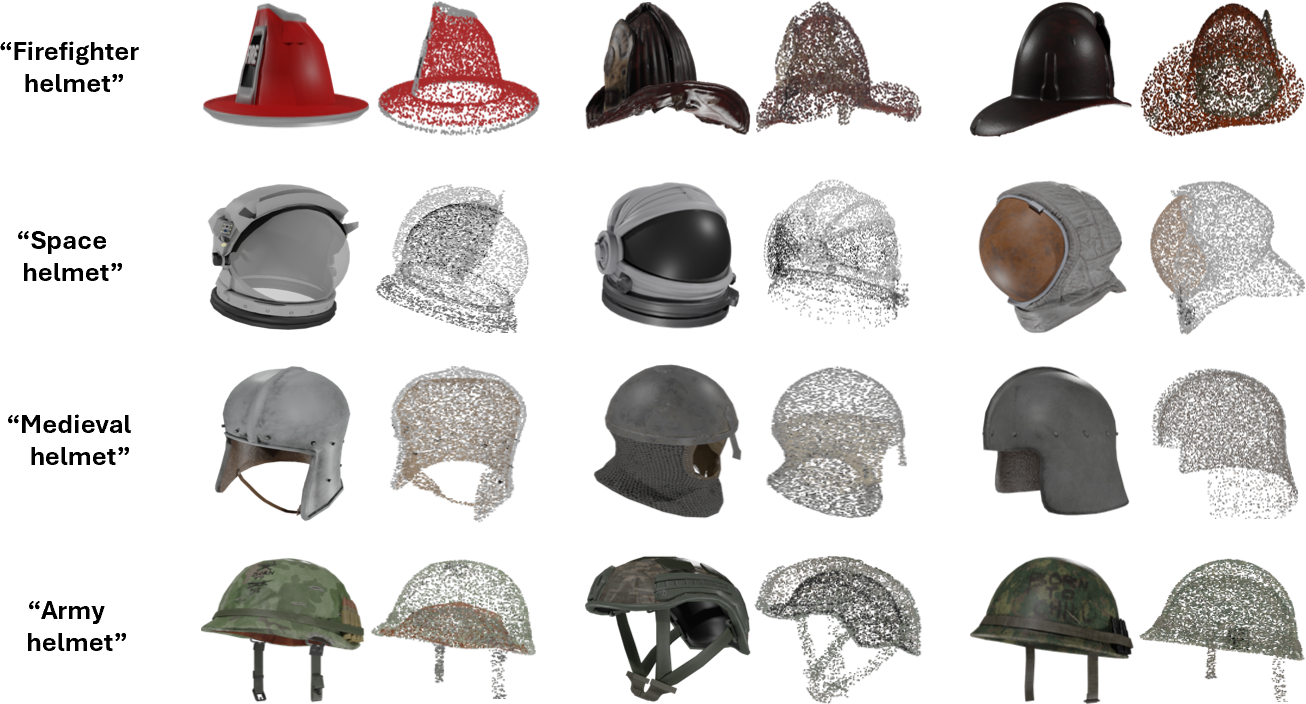} \\
    \hspace{0.02in} Input \hspace{0.5in} 1st Retrieved Result \hspace{0.4in} 2nd Retrieved Result \hspace{0.4in} 3rd Retrieved Result
    \caption{
    \textbf{Text-to-3D retrieval across closely related subcategories.}
    Each retrieved result is shown as a pair: an RGB rendering of the retrieved shape and its corresponding point cloud, which is the item returned by the search.
    These examples demonstrate fine-grained separability among "helmet" subcategories and strong text-to-3D cross-modal consistency.
    }
    \label{fig:txt_ret_helmets}
\end{figure*}

\begin{figure*}[tb]
    \centering
    \includegraphics[width=\linewidth,height=0.39\textheight]{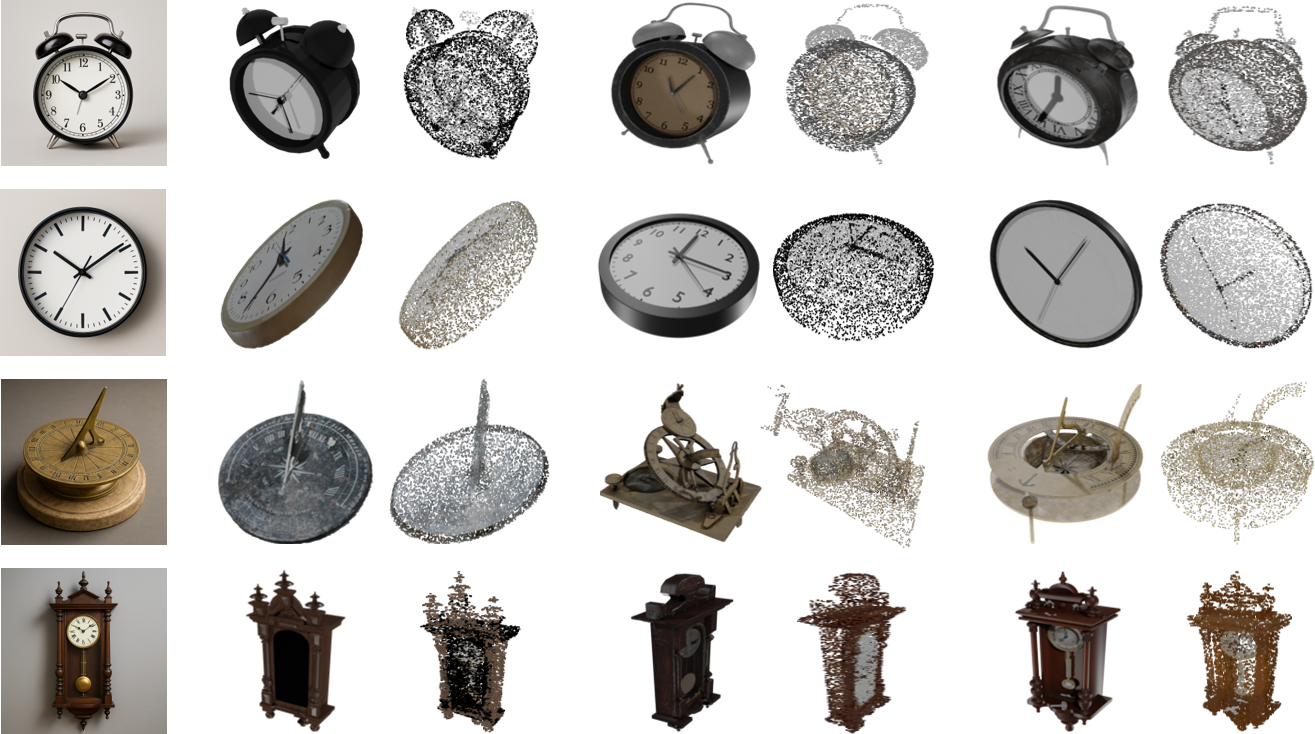} \\
    \hspace{0.05in} Input \hspace{0.485in} 1st Retrieved Result \hspace{0.375in} 2nd Retrieved Result \hspace{0.4in} 3rd Retrieved Result
    \caption{
    \textbf{Image-to-3D retrieval across closely related subcategories.}
    Each retrieved result is shown as a pair: an RGB rendering of the retrieved shape and its corresponding point cloud, which is the item returned by the search.
    These examples demonstrate fine-grained separability among "clock" subcategories and strong image-to-3D cross-modal consistency.
    }
    \label{fig:img_ret_clocks}
\end{figure*}

\begin{figure*}[h]
    \centering
    \includegraphics[width=\linewidth,height=0.395\textheight]{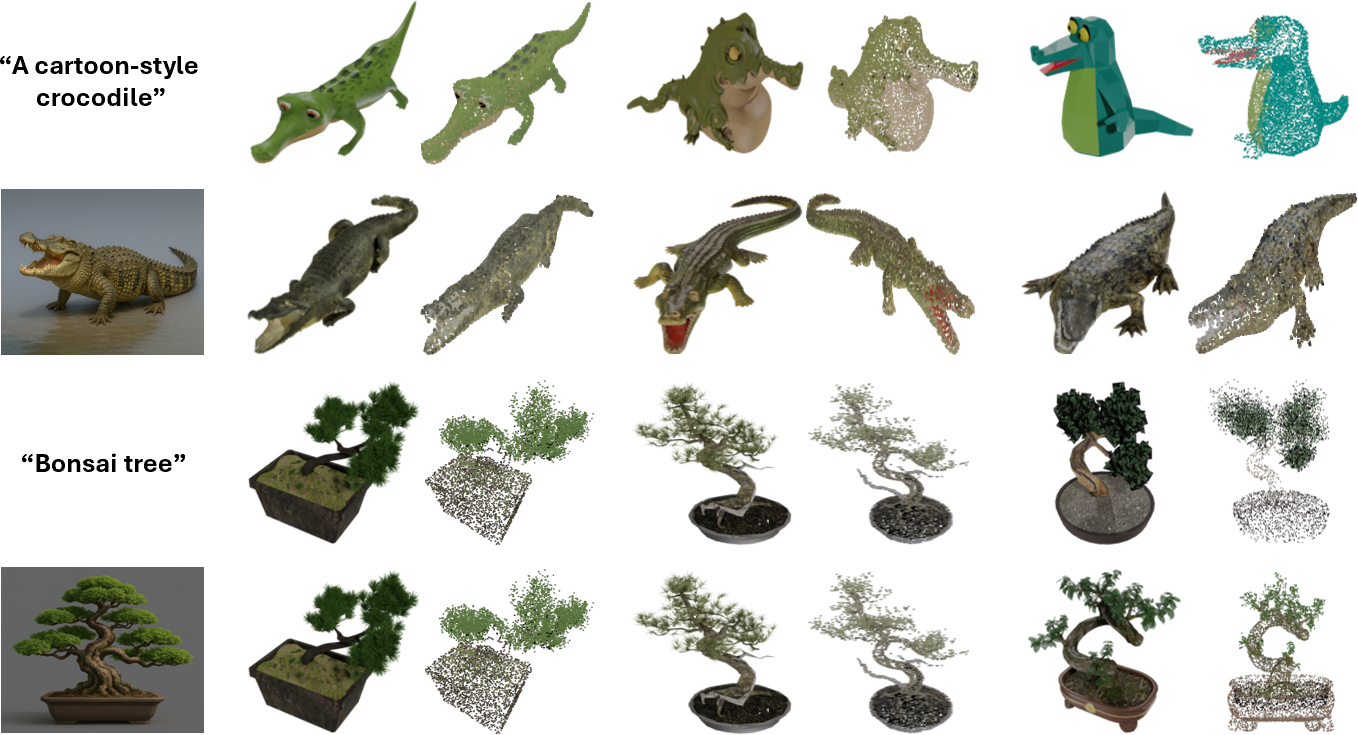} \\
    \hspace{0.15in} Input \hspace{0.525in} 1st Retrieved Result \hspace{0.4in} 2nd Retrieved Result \hspace{0.3in} 3rd Retrieved Result
    \caption{
    \textbf{Style-aware cross-modal retrieval.}
    Each retrieved result is shown as a pair: an RGB rendering of the retrieved shape and its corresponding point cloud, which is the item returned by the search.
    Top: style-mismatched text–image queries retrieve style-appropriate crocodile shapes, indicating the shared embedding space encodes semantic concepts across modalitites.
    Bottom: style-matched bonsai tree queries retrieve consistent bonsai shapes, showing robust alignment between modalities.
    }
    \label{fig:img_ret_croc_bonsai}
\end{figure*}

\noindent
\textbf{Cross-modal retrieval.}
Similarly to text-based retrieval, \cref{fig:teaser_image_3d_ret,fig:ret_img_head_cat,fig:ret_img_tail_cat} demonstrate that our method enables successful image-to-3D retrieval for both common (head) and rare (tail) categories.
Moreover, as illustrated in \cref{fig:txt_ret_helmets,fig:img_ret_clocks}, text and image queries from closely related subcategories retrieve the intended point clouds rather than visually similar distractors, demonstrating fine-grained separability and cross-modal consistency.
Finally, \cref{fig:img_ret_croc_bonsai} demonstrates faithful cross-modal alignment: semantically similar image and text queries (\eg, bonsai tree) retrieve nearly the same 3D shapes. 
The model is also style-aware: prompts like “A cartoon-style crocodile” return cartoon like crocodiles, whereas realistic photos retrieve realistic counterparts, showing that the shared embedding space encodes semantic concepts (\eg, style) across modalities.

\FloatBarrier
\clearpage

\section{Analysis of Hardness-Aware Regimes}
\label{supp_sec:analysis_of_hardness_aware_simple_loss}
To study how hardness-aware weighting affects training efficiency, we consider a simplified contrastive objective formulated as a linear combination of similarities between an anchor and its positive and negative keys. 
Specifically, we define:

{
\begin{equation}
\mathcal{L}^{\textit{Simple}}_{q_i}
=
\dfrac{1}{\tau}
\left[-\dfrac{1}{|P(i)|}\sum\limits_{r\in P(i)}s^{\scriptscriptstyle +}_{ir} 
\;+\; 
\sum\limits_{j\in N(i)} c_{ij}s^{\scriptscriptstyle -}_{ij} \right]\;\;.
\end{equation}
}

The positive-alignment component is adopted from the \emph{decoupled multi-positive contrastive loss} $\mathcal{L}^{\textit{DMP}}_{q_i}$, allowing us to isolate the effect of the negative-separation term and its hardness-aware weighting. 
In particular, the contribution of each negative similarity is modulated by the coefficient $c_{ij}$, which controls how strongly anchor $q_i$ is repelled from the negative key $j$.
Using \cref{eq:gradients,eq:prob_of_neg_samples,eq:pos_neg_gradients_relationship,eq:decoupled_negative_gradients} we define the following three sets of coefficients:

{
\begin{equation}
\renewcommand{\arraystretch}{3.0} 
c_{ij} =
\left\{
\begin{array}{l c} 
\text{without hardness-aware:}
&
\displaystyle 
\frac{1}{|N(i)|}\sum\limits_{k \in N(i)}
\frac{\partial \mathcal{L}^{\textit{MP}}_{q_i}}{\partial s^{\scriptscriptstyle -}_{ik}}
\\
\text{regular hardness-aware:}
&
\displaystyle 
\frac{\partial \mathcal{L}^{\textit{MP}}_{q_i}}{\partial s^{\scriptscriptstyle -}_{ij}}
\\
\text{amplified hardness-aware:}
&
\displaystyle 
\frac{\partial \mathcal{L}^{\textit{DMP}}_{q_i}}{\partial s^{\scriptscriptstyle -}_{ij}}
\end{array}
\right.
\end{equation}
}
We consider three choices for the coefficients $c_{ij}$ that progressively introduce the hardness-aware property into the negative-separation term.
\emph{Without hardness-aware weighting}, $c_{ij}$ is the same for all negatives associated with anchor $q_i$ (implemented as the mean gradient magnitude over negative samples under the standard multi-positive loss), so no negative sample is prioritized and all contribute equally. 
\emph{Regular hardness-aware weighting} sets $c_{ij}=\partial \mathcal{L}^{\textit{MP}}_{q_i}/\partial s^{\scriptscriptstyle -}_{ij}$, so negatives are weighted by their gradients under the standard multi-positive loss; since these gradients increase with higher similarity scores, harder (more confusable) negatives receive stronger repulsion, but the weighting remains coupled to positives through the shared softmax denominator.
\emph{Amplified hardness-aware weighting} instead uses $c_{ij}=\partial \mathcal{L}^{\textit{DMP}}_{q_i}/\partial s^{\scriptscriptstyle -}_{ij}$, where the normalization is over negatives only; consequently, negatives compete only with other negatives for weight, preserving and typically amplifying the emphasis on the hardest negatives even as the number of positive views grows.

As shown in \Cref{tab:influence_of_hardness_aware_in_simple_cl}, replacing uniformly weighted coefficients (denoted "without") with "regular" hardness-aware weighting improves Objaverse-LVIS Top-1/3/5 across all view counts. 
However, the improvement diminishes as more views are added; for Objaverse-LVIS Top-1, the gain drops from $0.8\%$ with a single view to $0.2\%$ with 12 views. 
In contrast, \Cref{tab:influence_of_hardness_aware_in_simple_cl} further shows that replacing the "regular" coefficients with the "amplified" variant yields substantial improvements at every view count, and these gains increase with more views, rising from $2.8\%$ at 1 view to $3.5\%$ at 12 views on Objaverse-LVIS Top-1. 
Together, these results confirm that hardness-aware weighting is crucial for effective optimization, and that coefficients derived from the \emph{decoupled multi-positive contrastive loss} further reinforce this effect by making negatives compete only with other negatives, thereby preventing gradient dilution as additional positives are introduced. 
Hence, decoupling amplifies the hardness-aware property and becomes progressively more beneficial in the multi-view regime.

\begin{table*}[tb]
\caption{{\bf Effect of hardness-aware coefficients in a simple contrastive learning objective.} 
The left-middle columns show that replacing equally weighted coefficients with “regular” hardness-aware weighting (as in the standard multi-positive formulation) improves Objaverse-LVIS Top-1/3/5 across all view counts. 
However, the margin shrinks as views increase (e.g., gains at 1 view are greater than gains at 12 views), reflecting the crowding of positives in the softmax. 
In contrast, the middle-right columns indicate that replacing “regular” hardness-aware coefficients with the “amplified” variant (derived from the decoupled multi-positive objective) yields the best accuracy at every view count, and crucially, the gains increase as the number of views grows.
Thus, decoupling amplifies the hardness-aware property and is especially beneficial in the multi-view regime. 
The experiment was conducted under the "ShapeNet" training regime without the text adapter.
}
\label{tab:influence_of_hardness_aware_in_simple_cl}
\centering
\begin{adjustbox}{max width=\textwidth}
\setlength{\tabcolsep}{3pt}
\begin{tabular}{@{}c | c c c | c c c | c c c@{}}
\toprule
\multicolumn{1}{c|}{\multirow{2}{*}{\textbf{\#Views}}} &
\multicolumn{3}{c|}{\textbf{Objaverse-LVIS Top-1}} &
\multicolumn{3}{c|}{\textbf{Objaverse-LVIS Top-3}} &
\multicolumn{3}{c}{\textbf{Objaverse-LVIS Top-5}} \\
& 
\textbf{Without} & \textbf{Regular} & \textbf{Amplified} &
\textbf{Without} & \textbf{Regular} & \textbf{Amplified} &
\textbf{Without} & \textbf{Regular} & \textbf{Amplified} \\
\midrule
1  & 12.1 & 12.9 & 15.7 & 20.5 & 21.8 & 26.9 & 24.8 & 26.2 & 32.8 \\
4  & 12.1 & 12.7 & 15.9 & 20.4 & 21.5 & 27.3 & 24.6 & 26.0 & 33.4 \\
8  & 12.6 & 13.0 & 16.2 & 21.1 & 21.9 & 27.7 & 25.2 & 26.2 & 33.6 \\
12 & 12.6 & 12.8 & 16.3 & 21.0 & 21.6 & 27.8 & 25.2 & 26.0 & 33.6 \\
\bottomrule
\end{tabular}
\end{adjustbox}
\end{table*}

\section{Further Ablation Study}
\label{supp_sec:additional_ablations}

As previously stated, for the ablation studies, we report results for a 32M-parameter PointBERT backbone trained on the "ShapeNet" training set, under the default setting of eight views for both the image and text modalities, unless stated otherwise.

\noindent
\textbf{Ablation on the temperature parameter $\tau$.}
We conduct an ablation study on the temperature parameter $\tau$ under the "ShapeNet" pretraining regime. 
While prior methods typically use a learnable temperature, often in the range of $0.07$ to $0.01$, \Cref{tab:constant_temperature} shows that, for the long-tail Objaverse-LVIS Top-1 metric, performance improves as the temperature decreases. 
In particular, the best accuracy is achieved at $\tau=0.01$. 
This behavior is expected, as a lower temperature sharpens the softmax, concentrating probability mass, and therefore gradient magnitude, on the hardest few negatives in the anchor’s neighborhood.

\begin{table}[tb]
\caption{
{\bf Ablation on the temperature parameter $\tau$}.
Accuracy improves as the temperature decreases, reaching its peak at $\tau=0.01$.
}
\label{tab:constant_temperature}
\centering
\small
\begin{adjustbox}{max width=\columnwidth}
\begin{tabular}{@{}c | c c c c c c c@{}}
\toprule
\textbf{Temperature Value $\tau$} & \textbf{learnable} & \textbf{0.1} & \textbf{0.07} & \textbf{0.04} & \textbf{0.02} & \textbf{0.01} & \textbf{0.005} \\
\midrule
\textbf{Objaverse-LVIS Top-1} & 17.3 & 14.0 & 14.7 & 16.0 & 17.1 & \textbf{17.7} & 16.5 \\
\bottomrule
\end{tabular}
\end{adjustbox}
\end{table}

\begin{table*}[tb]
\caption{
{\bf Batch size comparison for MP vs. DMP.} 
Zero-shot Top-1 accuracy comparison between the naive multi-positive (MP) baseline and our decoupled multi-positive (DMP) formulation across different batch sizes. 
Under the "Ensembled" pretraining regime, DMP yields substantial gains over MP on Objaverse-LVIS and ScanObjectNN at all batch sizes, while on ModelNet40 it is slightly worse for batch sizes above 512. 
All models are trained using a 26M-parameter PointBERT backbone without the text adapter.
For batch sizes 2048 and 4096, we use 10 and 30 warmup epochs, respectively, to account for the higher learning rate and the reduced number of iterations per epoch. 
}
\centering
\label{tab:zero_shot_bs_comp_mp_vs_dmp}
\begin{subtable}[t]{0.32\textwidth}
\centering
\caption{{\bf Objaverse-LVIS Top-1}}
\label{tab:zero_shot_bs_comp_mp_vs_dmp_o_lvis}
\begin{adjustbox}{max width=\textwidth}
\setlength{\tabcolsep}{3pt}
\begin{tabular}{@{}c c c c c c@{}}
\toprule
&
\multicolumn{5}{c}
{\textbf{Batch Size}} \\
& 256 & 512 & 1024 & 2048 & 4096 \\
\midrule
\textbf{MP} & 49.9 & 51.2 & 52.4 & 53.2 & 53.4 \\
\textbf{DMP} & \textbf{53.9} & \textbf{54.8} & \textbf{55.4} & \textbf{55.7} & \textbf{55.0} \\
\bottomrule
\end{tabular}
\end{adjustbox}
\end{subtable}
\hfill
\begin{subtable}[t]{0.32\textwidth}
\centering
\caption{{\bf ScanObjectNN Top-1}}
\label{tab:zero_shot_bs_comp_mp_vs_dmp_sonn}
\begin{adjustbox}{max width=\textwidth}
\setlength{\tabcolsep}{3pt}
\begin{tabular}{@{}c c c c c c@{}}
\toprule
&
\multicolumn{5}{c}
{\textbf{Batch Size}} \\
& 256 & 512 & 1024 & 2048 & 4096 \\
\midrule
\textbf{MP} & 57.3 & 58.7 & 60.4 & 60.2 & 60.3 \\
\textbf{DMP} & \textbf{63.8} & \textbf{63.8} & \textbf{64.2} & \textbf{64.6} & \textbf{64.0} \\
\bottomrule
\end{tabular}
\end{adjustbox}
\end{subtable}
\hfill
\begin{subtable}[t]{0.32\textwidth}
\centering
\caption{{\bf ModelNet40 Top-1}}
\label{tab:zero_shot_bs_comp_mp_vs_dmp_mnet40}
\begin{adjustbox}{max width=\textwidth}
\setlength{\tabcolsep}{3pt}
\begin{tabular}{@{}c c c c c c@{}}
\toprule
&
\multicolumn{5}{c}
{\textbf{Batch Size}} \\
& 256 & 512 & 1024 & 2048 & 4096 \\
\midrule
\textbf{MP} & 87.3 & \textbf{87.5} & \textbf{88.0} & \textbf{87.4} & \textbf{87.5} \\
\textbf{DMP} & \textbf{87.9} & 87.4 & 87.2 & 86.6 & 86.8 \\
\bottomrule
\end{tabular}
\end{adjustbox}
\end{subtable}
\end{table*}

\begin{table*}[tb]
\caption{{\bf Zero-shot shape classification comparison for different batch sizes.}
Under the "Ensembled" pretraining regime, higher batch sizes (1024 and 2048) improve Top-1 accuracy on Objaverse-LVIS, whereas smaller batch sizes (256 and 512) yield better Top-1 accuracy on ScanObjectNN and ModelNet40. 
Among all settings, batch size 512 provides the strongest overall trade-off, achieving the best performance on most metrics while remaining competitive elsewhere.
}
\label{tab:optimal_batch_size_selection}
\centering
\small
\begin{adjustbox}{max width=\textwidth}
\setlength{\tabcolsep}{6pt}
\begin{tabular}{@{}c | c c c | c c c | c c c@{}}
\toprule
\multicolumn{1}{c|}{\multirow{2}{*}{\textbf{Batch Size}}} &
\multicolumn{3}{c|}{\textbf{Objaverse-LVIS}} &
\multicolumn{3}{c|}{\textbf{ScanObjectNN}} &
\multicolumn{3}{c}{\textbf{ModelNet40}} \\
& 
\textbf{Top-1} & \textbf{Top-3} & \textbf{Top-5} &
\textbf{Top-1} & \textbf{Top-3} & \textbf{Top-5} &
\textbf{Top-1} & \textbf{Top-3} & \textbf{Top-5}\\
\midrule
256 & 54.7 & 77.1 & 83.4 & 65.7 & 84.0 & 91.0 & \textbf{87.8} & \textbf{97.0} & 98.3 \\
512 & 55.5 & \textbf{77.3} & \textbf{83.6} & \textbf{66.7} & \textbf{85.3} & \textbf{92.2} & 87.5 & 96.6 & 98.1 \\
1024 & \textbf{56.0} & 77.1 & 83.3 & 65.1 & 84.5 & 91.7 & 86.6 & \textbf{97.0} & \textbf{98.7} \\
2048 & 55.7 & 76.5 & 82.5 & 65.4 & 84.4 & 91.8 & 86.6 & 96.9 & 98.3 \\
\bottomrule
\end{tabular}
\end{adjustbox}
\end{table*}

\noindent
\textbf{Effect of batch size scaling.}
\Cref{tab:zero_shot_bs_comp_mp_vs_dmp_o_lvis} shows that DMP consistently outperforms the naive MP baseline across all batch sizes and continues to improve with increasing batch size, except at the largest batch size, on the Objaverse-LVIS Top-1 metric, demonstrating strong effectiveness even in low-resource regimes while still benefiting from larger batches when available. 
Replacing the naive multi-positive (or single-positive) loss with our decoupled formulation improves Top-1 accuracy by $1.6\%-4.0\%$ on Objaverse-LVIS and by $3.7\%-6.5\%$ on ScanObjectNN across all batch sizes, as shown in \Cref{tab:zero_shot_bs_comp_mp_vs_dmp_o_lvis,tab:zero_shot_bs_comp_mp_vs_dmp_sonn}. 
Notably, for both benchmarks, the best result achieved with MP remains below the worst result achieved with DMP, when the latter is obtained using the smallest batch size. 
On ModelNet40, as observed in \Cref{tab:zero_shot_bs_comp_mp_vs_dmp_mnet40}, DMP is superior or competitive at smaller batch sizes, while trailing naive MP by only up to $0.8\%$ at larger batch sizes.

In \Cref{tab:optimal_batch_size_selection}, we further analyze the effect of batch size on the full version of our method, including the text adapter and the weighting function for the multi-positive term. 
From \Cref{tab:zero_shot_bs_comp_mp_vs_dmp,tab:optimal_batch_size_selection}, we observe that these additions improve Top-1 accuracy on Objaverse-LVIS and ScanObjectNN compared with the variant that omits them, while having little effect on ModelNet40 Top-1 accuracy. 
We also observe from \Cref{tab:optimal_batch_size_selection} that larger batch sizes (1024 and 2048) are more effective for Objaverse-LVIS, whereas smaller batch sizes (256 and 512) yield better results on ScanObjectNN and ModelNet40. 
Among all settings, batch size 512 provides the strongest overall trade-off, delivering the best performance on most metrics while remaining competitive elsewhere.

For the batch size scaling experiments, we adopt a 26M-parameter PointBERT backbone, as it is the largest model that remains feasible to train with large batch sizes under our computational constraints. 
All experiments are conducted under the "Ensembled" pretraining regime, with a per-GPU batch size of up to 512; this accounts for the discrepancy in the batch-size-512 results reported in \Cref{tab:zero_shot_comp_res,tab:zero_shot_comp_res_pointbert_model_sizes}.

\begin{table*}[tb]
\caption{{\bf Zero-shot shape classification comparison for different PointBERT model sizes.} 
With the PointBERT backbone, HOLA exhibits consistent gains across all benchmarks as model capacity increases. 
The most important benchmark is Objaverse-LVIS, a very large dataset with a pronounced long-tail distribution, which is the main challenge our approach aims to address. 
}
\label{tab:zero_shot_comp_res_pointbert_model_sizes}
\centering
\begin{adjustbox}{max width=\textwidth}
\setlength{\tabcolsep}{3pt}
\begin{tabular}{@{}l c | c | c | c c c | c c c | c c c@{}}
\toprule
\multicolumn{1}{c}{\multirow{2}{*}
{\textbf{Method}}} &
\multicolumn{1}{c|}{\multirow{2}{*}{\textbf{\#Params $\downarrow$}}} &
\multicolumn{1}{c|}{\multirow{2}{*}{\textbf{FLOPs $\downarrow$}}} &
\multicolumn{1}{c|}{\multirow{2}{*}{\textbf{FPS $\uparrow$}}} &
\multicolumn{3}{c|}{{\textbf{Objaverse-LVIS}}} &
\multicolumn{3}{c|}{\textbf{ScanObjectNN}} &
\multicolumn{3}{c}{\textbf{ModelNet40}} \\
& & & &
\textbf{Top-1} & \textbf{Top-3} & \textbf{Top-5} &
\textbf{Top-1} & \textbf{Top-3} & \textbf{Top-5} &
\textbf{Top-1} & \textbf{Top-3} & \textbf{Top-5} \\
\midrule
\multicolumn{13}{c}{\textit{Trained on "Ensembled no LVIS" (829,460 triplets)}} \\
\midrule
\multirow{5}{*}{\textbf{HOLA}}
& 5.1M & 1G & 369 & 45.3 & 66.6 & 73.9 & 60.7 & 80.8 & 88.7 & 86.0 & 95.7 & 97.2 \\
& 13.3M & 2G & 359 & 46.9 & 68.4 & 75.5 & 61.7 & 81.6 & 89.9 & 86.3 & 96.0 & 97.6 \\
& 26.0M & 7G & 264 & 48.8 & 70.2 & 76.9 & 64.9 & 83.4 & 90.3 & 87.2 & 96.6 & 98.1 \\
& 32.3M & 29G & 202 & 50.3 & 71.4 & 78.0 & 67.2 & \textbf{85.4} & 91.9 & 88.3 & \textbf{97.6} & 98.9 \\
& 72.1M & 84G & 152 & \textbf{50.7} & \textbf{72.1} & \textbf{78.9} & \textbf{67.7} & \textbf{85.4} & \textbf{92.4} & \textbf{89.0} & \textbf{97.6} & \textbf{99.0} \\
\midrule
\multicolumn{13}{c}{\textit{Trained on "Ensembled" (875,665 triplets)}} \\
\midrule
\multirow{5}{*}{\textbf{HOLA}}
& 5.1M & 1G & 369 & 50.9 & 72.1 & 79.0 & 59.8 & 79.6 & 88.3 & 85.7 & 95.8 & 97.5 \\
& 13.3M & 2G & 359 & 53.5 & 75.0 & 81.3 & 63.2 & 82.5 & 90.5 & 86.5 & 96.1 & 97.6 \\
& 26.0M & 7G & 264 & 55.7 & 77.3 & 83.6 & 65.1 & 83.7 & 90.9 & 87.5 & 97.4 & 98.7 \\
& 32.3M & 29G & 202 & 56.7 & 78.4 & 84.4 & 68.3 & \textbf{86.7} & 93.0 & 88.1 & 97.4 & 98.6 \\
& 72.1M & 84G & 152 & \textbf{57.3} & \textbf{79.0} & \textbf{84.9} & \textbf{68.7} & 86.5 & \textbf{93.1} & \textbf{88.8} & \textbf{97.6} & \textbf{98.8} \\
\bottomrule
\end{tabular}
\end{adjustbox}
\end{table*} 

\noindent
\textbf{Effect of model scaling.}
\Cref{tab:zero_shot_comp_res_pointbert_model_sizes} analyzes how scaling the PointBERT backbone affects zero-shot accuracy under the “Ensembled” and "Ensembled no LVIS" pretraining regimes. 
As model capacity increases from $5.1M$ to $72.1M$ parameters, performance improves consistently across the benchmarks. 
On the long-tail Objaverse-LVIS benchmark, Top-1 accuracy rises from $50.9\%$ to $57.3\%$ as model size scales under the "Ensembled" setting, with a similar trend observed under the "Ensembled no LVIS" setting, where performance improves from $45.3\%$ to $50.7\%$.
The same trend is observed on the real-scan ScanObjectNN benchmark, with Top-1 accuracy increasing from $59.8\%$ to $68.7\%$ for the "Ensembled" setting and from $60.7\%$ to $67.7\%$ for the "Ensembled no LVIS" setting.
These gains come with the expected compute trade-off: FLOPs grow from $1G$ to $84G$ and throughput decreases from 369 FPS to 152 FPS, highlighting a clear accuracy–efficiency scaling curve.

\begin{table*}[tb]
\caption{{\bf Extended ablation on the effect of the text adapter.} 
Introducing the text adapter consistently improves accuracy on most metrics. 
Combining it with an EMA-smoothed variant on the text-3D pathway and the image–text loss yields the largest gains, improving results across all benchmarks and metrics compared to not using it.}
\label{tab:extended_text_adapter_ablation}
\centering
\begin{adjustbox}{max width=\textwidth}
\setlength{\tabcolsep}{2pt}
\begin{tabular}{@{}c c c | c c c | c c c | c c c@{}}
\toprule
\multicolumn{1}{c}{\multirow{2}{*}{\shortstack[c]{\textbf{Text}\\\textbf{Adapter}}}} &
\multicolumn{1}{c}{\multirow{2}{*}{\shortstack[c]{\textbf{Image-Text}\\\textbf{Loss}}}} &
\multicolumn{1}{c|}{\multirow{2}{*}{\shortstack[c]{\textbf{EMA}\\(with stop-gradient)}}} &
\multicolumn{3}{c|}{{\textbf{Objaverse-LVIS}}} &
\multicolumn{3}{c|}{\textbf{ScanObjectNN}} &
\multicolumn{3}{c}{\textbf{ModelNet40}} \\
& & &
\textbf{Top-1} & \textbf{Top-3} & \textbf{Top-5} &
\textbf{Top-1} & \textbf{Top-3} & \textbf{Top-5} &
\textbf{Top-1} & \textbf{Top-3} & \textbf{Top-5}\\
\midrule
\xmark & \xmark & \xmark & 17.1 & 29.5 & 35.5 & 51.9 & 74.7 & 86.6 & 73.5 & 88.0 & 92.3 \\
\cmark & \xmark & \xmark & 17.3 & 29.7 & 35.6 & 55.8 & \textbf{78.2} & 87.5 & 73.8 & \textbf{91.2} & \textbf{95.1} \\
\cmark & \cmark & \xmark & 17.2 & 29.9 & 35.8 & \textbf{56.0} & 75.8 & 85.2 & 70.9 & 88.6 & 93.6 \\
\cmark & \cmark & \cmark & \textbf{17.7} & \textbf{30.0} & \textbf{36.1} & 55.9 & 77.4 & \textbf{87.7} & \textbf{74.2} & 90.0 & 95.0 \\
\bottomrule
\end{tabular}
\end{adjustbox}
\end{table*}

\noindent
\textbf{Extended ablation of the text adapter and related components.}
Our text supervision comes from heterogeneous sources, which are distributionally mismatched: in-domain annotations/VLM captions on one side and retrieved web captions on the other.
Although web captions provide complementary knowledge, a domain gap (\eg, style, multilingual) prevents fully exploiting it.
We therefore introduce a lightweight text adapter that narrows this gap by aligning web-caption embeddings with the in-domain image/text manifold, unlocking their value for language–3D distillation. 
Across configurations, the adapter is consistently beneficial, as reflected in \Cref{tab:extended_text_adapter_ablation}, with particularly strong gains for the Objaverse-LVIS and ScanObjectNN benchmarks.

Using only the text adapter with an image-text loss leaves the text-3D alignment vulnerable to target drift: because the adapter updates at every training step, the 3D encoder is optimized against continually moving text embeddings, effectively “chasing” a shifting target and impeding learning. 
To mitigate this, we maintain an exponential moving average (EMA) copy of the adapter for the text-3D pathway and use its outputs as teacher targets that change slowly across training steps, providing more stable supervision. 
The EMA parameters are updated exclusively by exponential averaging (no backpropagation), which improves optimization stability. 
Coupled with the image-text loss, this stabilization yields the largest gains, as shown in \Cref{tab:extended_text_adapter_ablation}, with the most pronounced improvements on the long-tail Objaverse-LVIS benchmark.

When the input to the text adapter is changed from web captions to in-domain annotations/VLM captions, Objaverse-LVIS Top-1 accuracy drops by $5.8\%$ under the "ShapeNet" pretraining regime. 
This indicates that the adapter’s main benefit lies in bridging the domain gap associated with web captions. 
Furthermore, replacing the text adapter with an image adapter for text–image alignment, as suggested in \cite{Zhang_2024_CVPR}, leads to a $3.5\%$ drop in accuracy, underscoring the importance of adapting the text modality in our framework.


\section{Implementation Details}
\label{supp_sec:implementation_details}

We follow \cite{NEURIPS2023_8c7304e7} and use OpenCLIP ViT-bigG-14 \cite{cherti2023reproducible} as the frozen CLIP teacher. 
We train all models and run ablation studies on a server with 8 NVIDIA A100 GPUs. 
For the "ShapeNet" setting, we train on a single GPU with a global batch size of 200. 
For the "Ensembled" and "Ensembled no LVIS" settings, we train using 4 GPUs with a per-GPU batch size of 128 (global batch size of 512).
We set the base learning rate to $5\times10^{-4}$ for "ShapeNet" and $6\times10^{-4}$ for "Ensembled" / "Ensembled no LVIS". 
We scale the learning rate with the number of views to reflect the increased effective batch size, up to caps of $1\times10^{-3}$ and $1.2\times10^{-3}$, respectively.
We further scale the learning rate to account for the batch size: $lr = base\_lr \cdot \max(1, \sqrt{batch\_size/256})$.
To stabilize and improve the evaluation accuracy, we maintain an exponential moving average (EMA) of the 3D encoder parameters with a decay rate of 0.9995, following \cite{tarvainen2017mean,Gao_2024_CVPR}.
Following prior work, we set $\lambda_1=\lambda_2=\lambda_3=1$.
HOLA is pretrained for 160 epochs using AdamW optimizer \cite{loshchilov2017decoupled} with a cosine learning rate scheduler \cite{loshchilov2016sgdr} and a 5-epoch linear warmup.
For PointBERT’s \cite{yu2022point} geometric operations (e.g., farthest point sampling), we use the efficient implementation from \cite{pytorchpointnet++}.

\end{document}